\documentclass[sigconf]{acmart}
\AtBeginDocument{%
  }

\setcopyright{cc}
\copyrightyear{2026}
\acmYear{2026}
\acmDOI{10.1145/3788646.3789533}
\acmConference[CSLAW '26]{Make sure to enter the correct
  conference title from your rights confirmation email}{March 3–5, 2026}{Berkeley, CA, USA}
\acmBooktitle{CS\&law '26: ACM Symposium on Computer Science and Law, March 03--05, 2026, Berkeley, CA, USA}
\acmISBN{979-8-4007-2447-3/26/03}

\acmYear{2026}\copyrightyear{2026}
\setcopyright{cc}
\setcctype[4.0]{by}
\acmConference[CSLAW '26]{Symposium on Computer Science and Law}{March 3--5, 2026}{Berkeley, CA, USA}
\acmBooktitle{Symposium on Computer Science and Law (CSLAW '26), March 3--5, 2026, Berkeley, CA, USA}
\acmDOI{10.1145/3788646.3789533}
\acmISBN{979-8-4007-2447-3/26/03}
\usepackage{svg} 
\usepackage{multirow}
\usepackage{listings}
\usepackage{float}

\begin{document}
\settopmatter{printacmref=true}
\title{Benchmarking Legal RAG: The Promise and Limits of AI Statutory Surveys}


\author{Mohamed Afane}
\email{afane@law.stanford.edu}
\affiliation{%
  \institution{Stanford University}
  \city{Stanford}
  \state{CA}
  \country{USA}}

\author{Emaan Hariri}
\email{ehariri@stanford.edu}
\affiliation{%
  \institution{Stanford University}
  \city{Stanford}
  \state{CA}
  \country{USA}}

\author{Derek Ouyang}
\email{douyang1@law.stanford.edu}
\affiliation{%
  \institution{Stanford University}
  \city{Stanford}
  \state{CA}
  \country{USA}}

\author{Daniel E. Ho}
\email{dho@law.stanford.edu}
\affiliation{%
  \institution{Stanford University}
  \city{Stanford}
  \state{CA}
  \country{USA}}
\renewcommand{\shortauthors}{Afane et al.}

\begin{abstract}
Retrieval-augmented generation (RAG) offers significant potential for legal AI, yet systematic benchmarks are sparse. Prior work introduced LaborBench to benchmark RAG models based on ostensible ground truth from an exhaustive, multi-month, manual enumeration of all U.S. state unemployment insurance requirements by U.S.\ Department of Labor (DOL) attorneys. That prior work found poor performance of standard RAG (70\% accuracy on Boolean tasks). Here, we assess three emerging tools not previously evaluated on LaborBench: the Statutory Research Assistant (STARA), a custom statutory research tool, and two commercial tools by Westlaw and LexisNexis marketing AI statutory survey capabilities. We make five main contributions. First, we show that STARA achieves substantial performance gains, boosting accuracy to 83\%. Second, we show that commercial platforms fare poorly, with accuracy of 58\% (Westlaw AI) and 64\% (Lexis+ AI), even worse than standard RAG. Third, we conduct a comprehensive error analysis, comparing our outputs to those compiled by DOL attorneys, and document both reasoning errors, such as confusion between related legal concepts and misinterpretation of statutory exceptions, and retrieval failures, where relevant statutory provisions are not captured. Fourth, we discover that many apparent errors are actually significant omissions by DOL attorneys themselves, such that STARA's actual accuracy is 92\%. Fifth, we chart the path forward for legal RAG through concrete design principles,  offering actionable guidance for building AI systems capable of accurate multi-jurisdictional legal research.
\end{abstract}




\maketitle

\section{Introduction}
\label{sec:intro}

A mainstay of U.S. and comparative legal research are statutory surveys \cite{morainStateLevelSupportTobacco2018,w.hahnStateFederalRegulatory2000,hamillCertainDeathFiftyState2007}. A common task for lawyers, policymakers, and researchers is to understand how legal requirements vary across jurisdictions. In unemployment insurance (UI), for instance, the U.S.\ Department of Labor (DOL) tracks over 101 distinct dimensions across all fifty states, each with tremendous consequences for workers and employers. This annual compilation of state UI laws represents a monumental undertaking, requiring teams of federal attorneys working over six months to systematically review and document statutory provisions across every state code, culminating in a 200-page publication of comparison tables \cite{haririAIStatutorySimplification2025}. The substantial manual effort expended and required for these surveys reflects both their importance and the inherent difficulty of conducting comprehensive, multi-jurisdictional statutory analysis.

Researchers have long explored the potential of computational tools to assist with legal reasoning and statutory interpretation \cite{lockeCaseLawRetrieval2022, ouelletteCanAIHold2025}. The emerging promise of artificial intelligence (AI) assistance for such systematic surveys has sparked significant interest, with commercial legal research platforms, such as Westlaw AI and Lexis+ AI, widely marketing their AI capabilities for fifty-state surveys across numerous areas of law \cite{50StateSurveys2025, AIPowered50State2025}. 

Yet statutory analysis remains a frontier challenge for AI systems \cite{guhaStateStatutesProject2024,hamillCertainDeathFiftyState2007}. The hierarchical structure of legal codes, extensive cross-references between provisions, precisely defined terms that differ from common usage, and complex interdependencies all create obstacles that standard natural language processing approaches may not adequately address. In a previous academic research effort involving authors of this study, \citet{suraniWhatLawSystem2025a} developed the Statutory Research Assistant (STARA), a specialized retrieval system leveraging domain-specific preprocessing and attention to statutory structure. 

What is particularly lacking is rigorous benchmarking to assess performance of different approaches to AI statutory surveys.  Responding to this gap, \citet{haririAIStatutorySimplification2025} introduced LaborBench, a benchmark for evaluating AI performance on state UI laws. Their evaluation of large language models (LLM) with retrieval-augmented generation (RAG) revealed poor performance on statutory questions, with even the most advanced models achieving F1 scores below 70\%. 
LaborBench's foundation in real DOL compilations makes it particularly compelling, as it reflects actual questions that require extensive manual effort by federal agency experts to answer. These findings highlighted that generic LLMs, despite broad capabilities in legal reasoning tasks \cite{guhaLegalBenchCollaborativelyBuilt2023b}, struggle with the specific demands of complex statutory analysis. \citet{haririAIStatutorySimplification2025}, however, did not evaluate recent models. 

We present the first systematic evaluation of STARA, Westlaw AI, and Lexis+ AI on LaborBench, including several contributions:

\begin{enumerate} 
\item \textbf{System Performance}: STARA achieves 83\% accuracy and 81\% F1 score on the LaborBench benchmark, outperforming the best models evaluated in the original LaborBench paper by 14\% in both accuracy and F1 score, demonstrating substantial improvements over current state-of-the-art approaches to statutory analysis.

\item \textbf{Commercial Platform Evaluation}: We conduct the first systematic evaluation of Westlaw AI and Lexis+ AI on LaborBench, which achieve F1 scores of 64\% and 41\%. In comparison, a baseline answering affirmatively for all questions achieves an F1 score of 73\%. We analyze apparent limitations in these widely adopted commercial systems including severe input context restrictions and systematic reasoning errors.

\item \textbf{Systematic Error Analysis}: We characterize the frontier challenges of legal RAG through comprehensive error analysis, identifying distinct failure modes in reasoning and retrieval. We also document persistent challenges across all evaluated AI systems, including confusion between related legal concepts and misinterpretation of statutory exceptions.

\item \textbf{DOL Compilation Gaps Discovery}: We show that many apparent errors are actually incorrect omissions by DOL attorneys themselves. STARA, for instance, identifies verifiable self-employment assistance programs in five states that were missed by DOL. This in turn boosts STARA's actual accuracy and F1 score to 92\% and 91\%. 

\item \textbf{Multi-Jurisdictional Survey Principles}: We establish concrete design principles for effective legal RAG systems based on our evaluation. These principles address the specific challenges of conducting systematic statutory analysis across state boundaries, offering actionable guidance for researchers and practitioners.

\end{enumerate}

The remainder of this paper proceeds as follows. Section~\ref{sec:background} provides background and related work. Section~\ref{sec:method} describes our experimental methodology and evaluation setup. Section~\ref{sec:results} presents results comparing system performance across multiple metrics and error categories. Section~\ref{sec:discussion} discusses implications for legal AI development and deployment, study limitations, and future research directions. Section~\ref{sec:conclusion} concludes.

\section{Background and Related Works}
\label{sec:background}

\subsection{Multi-Jurisdictional Statutory Analysis}
\label{sec:background-prior}

Multi-jurisdictional statutory analysis represents a cornerstone of legal research and policy evaluation. Researchers have long undertaken such comparative statutory surveys to understand regulatory variations and their impacts. \citet{morainStateLevelSupportTobacco2018} conducted comprehensive analysis across five states to examine tobacco control policies, demonstrating the methodological challenges inherent in cross-state legal research. Similarly, \citet{w.hahnStateFederalRegulatory2000} assessed regulatory reform initiatives across more than half of U.S. states, and \citet{hamillCertainDeathFiftyState2007} required extensive manual compilation to survey tax policies across all fifty states, highlighting the substantial resources required for comprehensive cross-state analysis. More recently, \citet{guhaStateStatutesProject2024} identified the absence of readily accessible databases for empirical research on state statutes, noting that current systems fail to support fine-grained statutory research and cannot effectively track trends in statutory adoption across jurisdictions. \citet{zheng2025reasoning} introduced reasoning-focused legal retrieval benchmarks for housing statutes, demonstrating that legal retrieval tasks requiring substantial reasoning between queries and relevant passages pose challenges for standard retrieval methods. These challenges are particularly acute in labor law, where nationwide data reveal systemic failures in core protections and widespread violations of statutory requirements \cite{BrokenLawsUnprotected2009}, underscoring the need for accessible, systematic statutory comparison tools.
\subsection{Unemployment Insurance and LaborBench}
\label{sec:background-laborbench}

\begin{figure*}[!t]
    \centering
    \includegraphics[width=\linewidth]{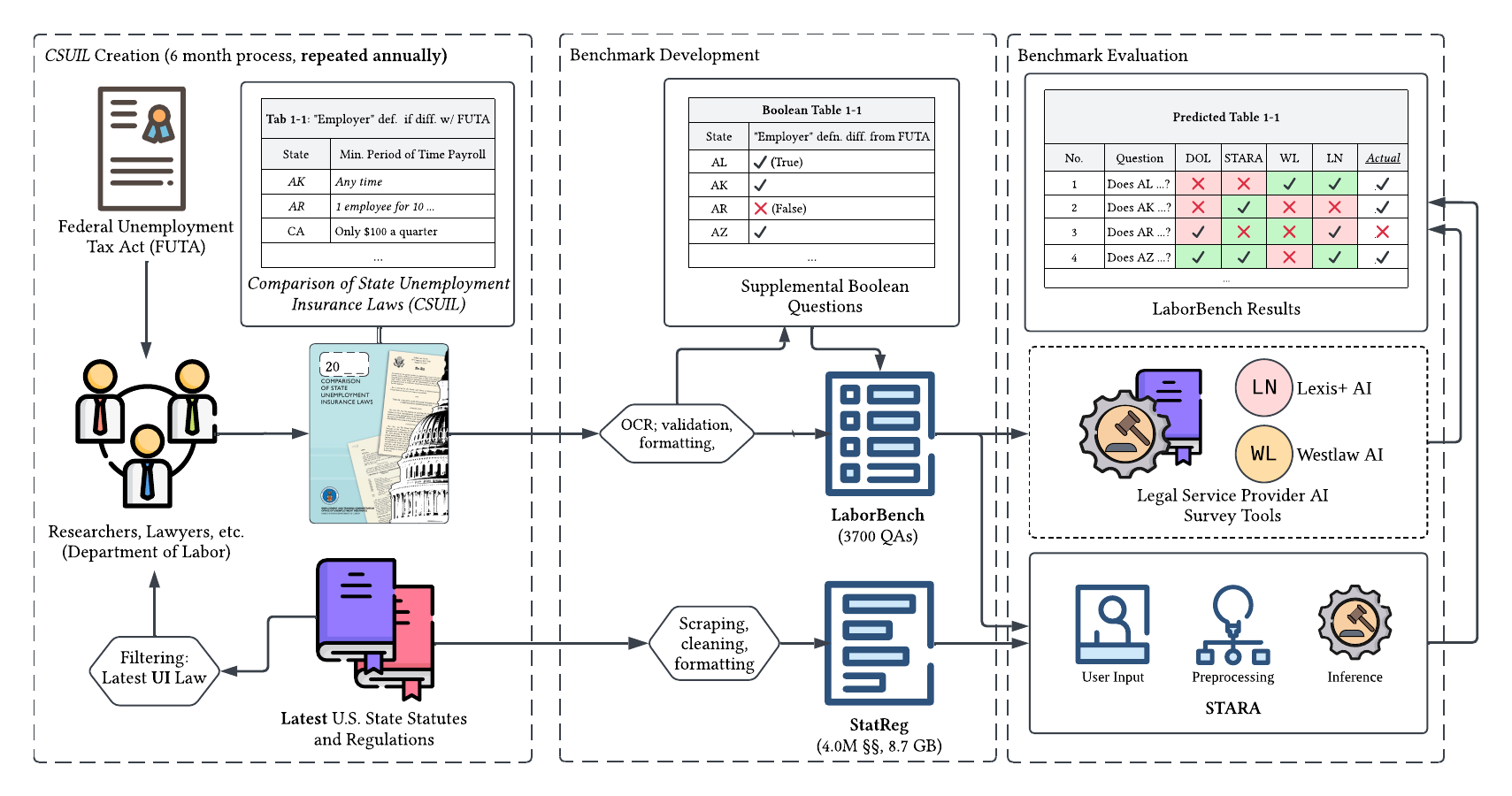}
    \caption{Summary of our benchmarking process. DOL = United States Department of Labor; UI = Unemployment insurance; OCR = Optical character recognition; QA = question/answer; STARA = Statutory Research Assistant.} \label{fig:splash}
\end{figure*}

Unemployment insurance is a highly consequential domain for multi-jurisdictional statutory research. UI programs function as the core federal-state social safety net provided to workers and communities at times of job instability, such as the COVID-19 pandemic when as many as 46 million Americans received unemployment payments \cite{stettner14Workers2021}. Amidst such high-stakes applications, UI requirements are also deeply complex, constantly changing, and jurisdictionally varied, complicating the ability of administrators, lawyers, policymakers, and researchers to monitor compliance and engage in reform efforts (see Appendix~\ref{app:reading-provisions}). This predicament, emblematic of U.S. benefits systems, is what leads Pahlka to devote an entire chapter of \textit{Recoding America} to the story of the ``new guy'', christened as such because he had \textit{only} been processing claims for 17 years compared to his more senior colleagues, yet still fundamentally learning the ropes \cite{pahlkaRecodingAmericaWhy2023a}. It is also what has fueled calls and efforts to leverage AI for both benefits administration and statutory simplification \cite{zeichnerIntroducingArtificialIntelligence2024,bainCuttingPolicyCruft2024}, which each depend on precise and comprehensive statutory interpretation. 

Responding to this important domain and to the challenges documented by prior scholarship in multi-jurisdictional statutory research, \citet{haririAIStatutorySimplification2025} developed LaborBench to benchmark AI performance on cross-state statutory analysis, specifically within UI law. While existing legal benchmarks have comprehensively evaluated AI performance across diverse legal tasks \cite{chalkidisLexGLUEBenchmarkDataset2022, feiLawBenchBenchmarkingLegal2023, guhaLegalBenchCollaborativelyBuilt2023b}, LaborBench focuses on the particular difficulty of multi-jurisdictional analysis, where practitioners must navigate fifty distinct statutory frameworks that address similar concepts through different provisions, definitions, and exceptions. The benchmark requires integration of information across multiple statutory sections, interpretation of defined terms in context, and reasoning about hierarchical relationships within legal codes. This type of statutory interpretation demands mastery of cross-references, scattered provisions, and amendments with complex effective dates. As Pahlka notes, such dense regulatory texts create immense barriers and difficulties for digital tools \cite{pahlkaRecodingAmericaWhy2023a}, which is evident in the benchmark's results: Claude, ChatGPT, Gemini, and other leading models all achieved F1 scores below 70\%, revealing the specific challenges that multi-jurisdictional statutory analysis poses even for models that perform well on other legal tasks.

\subsection{STARA and Domain-Specific Retrieval}
\label{sec:background-stara}

STARA is a specialized tool for conducting comprehensive statutory research across large legal codes. The system employs a multi-stage process to identify all provisions relevant to a user-defined legal question. It first parses and segments statutory text while preserving hierarchical structure, then augments provisions with necessary context including definitions, cross-references, and parent provisions. STARA applies user-specified criteria through optional keyword filtering (for computational efficiency) followed by language model classification to determine relevance. The tool was validated against human-compiled statutory surveys in multiple domains, reproducing them with high fidelity while uncovering additional relevant provisions that human researchers had missed. In formal evaluations, STARA achieved near–perfect recall and high precision on tasks such as enumerating federal criminal statutes and congressionally mandated reports, surfacing hundreds of provisions absent from the best available human datasets \cite{suraniWhatLawSystem2025a}. These results show that careful preprocessing and attention to statutory hierarchy can substantially outperform generic retrieval methods. While STARA achieved those results on single corpora (\textit{e.g.}, only the U.S.\ Code), it has not been assessed for exhaustive, fifty-state analysis like that required in LaborBench.

\subsection{Commercial Jurisdictional Survey Tools}
\label{sec:background-commercial}
Legal research platforms are increasingly promoted as AI solutions for legal studies and practice. However, independent studies show that the performance of AI systems on legal reasoning tasks remains uneven across areas of law \cite{dahlLargeLegalFictions2024b,mageshHallucinationFreeAssessingReliability2025b}. To date, over four hundred court cases worldwide have involved citations or statutes fabricated by AI tools, with the reported incidents largely arising from consumer-facing large language models rather than commercial legal research platforms \cite{AILawTracker2025, AIHallucinationCases2025}. Despite these concerns, legal service providers have specifically marketed new functions for AI multi-jurisdictional surveys. Westlaw advertises the ability to “compare statutes and regulations from all states with one easy search,” promising to “save hours or even days by searching all 50 states at once” and delivering “comprehensive reports” with “current and thorough findings” across varied state language and numbering systems \cite{50StateSurveys2025}. Lexis+ AI touts their “groundbreaking” AI-powered fifty-state surveys as transforming what “traditionally required weeks of painstaking research” into a process completed “in minutes.” Although few technical details are provided, LexisNexis describes its system as automatically identifying, comparing, and summarizing laws across all federal and state statutes and administrative codes on any topic, proclaiming “the implications for legal practice are substantial” \cite{AIPowered50State2025}. Both platforms emphasize dramatic time savings and accuracy, asserting that their technology reduces the risk of missing relevant provisions and enables users to find everything on their topic. Despite these bold marketing claims,  rigorous benchmarking is needed to verify whether these systems can deliver the promised accuracy and completeness.

\section{Methodology}
\label{sec:method}

\subsection{Experimental Setup}
\label{sec:method-setup}

Following the framework established in LaborBench \cite{haririAIStatutorySimplification2025}, we focus on binary classification tasks to enable systematic validation of our approach. This evaluation includes 1{,}647 questions on complex statutory UI laws, covering employment-related program availability, benefits eligibility requirements, calculation methods for UI claims, and other intricate aspects of UI frameworks. These questions ask whether specific states have particular laws or use certain legal provisions, requiring deep understanding of statutory text and cross-jurisdictional variations. The steps in our benchmarking pipeline are outlined in Figure \ref{fig:splash}. For our evaluation of STARA, we ran the system across full state UI codes to test large-scale statutory retrieval. 

An underlying challenge to applying generative AI to all state statutes are compute costs. STARA addresses this by allowing for optional regular expression filters (RegEx) to narrow the search to a subset of relevant provisions, a set still too large for manual review but well suited to STARA's semantic reasoning. However, these filters can also inadvertently exclude valid provisions, trading off computational speed and completeness. We applied RegEx filters to focus on UI law provisions across all 50 states. The benchmark contains 40 different question types, divided into 8 batches of 5 questions each for processing by STARA, with each batch using one common RegEx filter tailored to those specific question types (examples provided in the Appendix~\ref{app:regex}). The system processes complete state UI codes, maintains hierarchical structure and cross-reference relationships essential for accurate legal reasoning, and generates retrieved passages that are then processed to produce binary classifications with supporting reasoning and direct statutory citations.

\subsection{Commercial Platform Evaluation}
\label{sec:method-commercial}

We evaluate two specific AI multi-jurisdictional survey tools in commercial legal AI platforms. In Lexis+ AI, the  Protege tool offers two evaluation modes: users can either select up to three specific jurisdictions for targeted analysis or conduct what LexisNexis markets as a ``full survey'' across all states. The platform's 5{,}000-character limit allowed us to provide full context and questions for each of the 40 question types without modification. For our systematic evaluation, we utilized the full survey option to assess performance across all jurisdictions simultaneously, ensuring comprehensive coverage rather than selective state-by-state testing. Our initial evaluations revealed that answers sometimes differ between the targeted jurisdiction and full survey approaches, though overall accuracy remains comparable. Section~\ref{sec:results-commercial} provides detailed analysis of these variations and their implications for system reliability. In Westlaw AI, we focus on the AI Jurisdictional Surveys tool, which imposes a 300-character limit on query inputs. While the questions themselves remained identical across all systems, we had to substantially condense the contextual information that typically accompanies each question to fit within this constraint (see Appendix~\ref{app:input-examples}).

\subsection{Validation of System Outputs and DOL Report Accuracy}
\label{sec:method-dol}

After benchmarking the three systems, we conduct a detailed error analysis to understand the reasons for errors (\textit{e.g.}, retrieval, reasoning, classification) and to develop a nuanced understanding of the trajectory of capacities. 
This analysis began with a representative subset of LaborBench questions and involved close review of the underlying state labor statutes against the DOL report to understand error patterns.
A substantial share of apparent false positives proved to be valid statutory provisions that were simply absent from the DOL compilation. We thus proceeded with a more comprehensive validation process to separate omissions in the DOL compilation from classification mistakes, though, given resource limits and the much higher error volume from commercial systems, these corrections focused on STARA’s apparent false positives and false negatives (see Section~\ref{sec:results-error-summary}). 

\section{Results}
\label{sec:results}

\subsection{Overall Performance Comparison}
\label{sec:results-overall}

\begin{figure*}[t]
    \centering
    \begin{minipage}[t]{0.5\textwidth}
        \centering
        \vspace*{-29ex}
        \begin{tabular}[t]{|p{1.5cm}|c|c|c|c|}
            \hline
            \textbf{System} & \textbf{Accuracy} & \textbf{Precision} & \textbf{Recall} & \textbf{F1} \\
            \hline
            Baseline & 0.50 & 0.50 & \textbf{1.00} & 0.67 \\
            RAG & 0.66 & 0.57 & 0.81 & 0.67 \\
            Westlaw AI & 0.58 & 0.50 & 0.91 & 0.64 \\
            Lexis+ AI & 0.64 & 0.69 & 0.29 & 0.41 \\
            STARA & 0.83 & 0.76 & 0.87 & 0.81 \\
            \hline
            STARA (Corrected) & \textbf{0.92} & \textbf{0.94} & 0.89 & \textbf{0.91} \\
            \hline
        \end{tabular}
        \captionof{table}{Performance Comparison across AI systems. The baseline represents a majority class classifier. RAG represents the best performing retrieval-augmented generation model tested by \citet{haririAIStatutorySimplification2025}. STARA (Corrected) shows performance after accounting for provisions missed in DOL compilation.}
        \label{tab:performance_comparison}
    \end{minipage}%
    \hfill%
    \begin{minipage}[t]{0.45\textwidth}
        \centering
        \includegraphics[width=\linewidth]{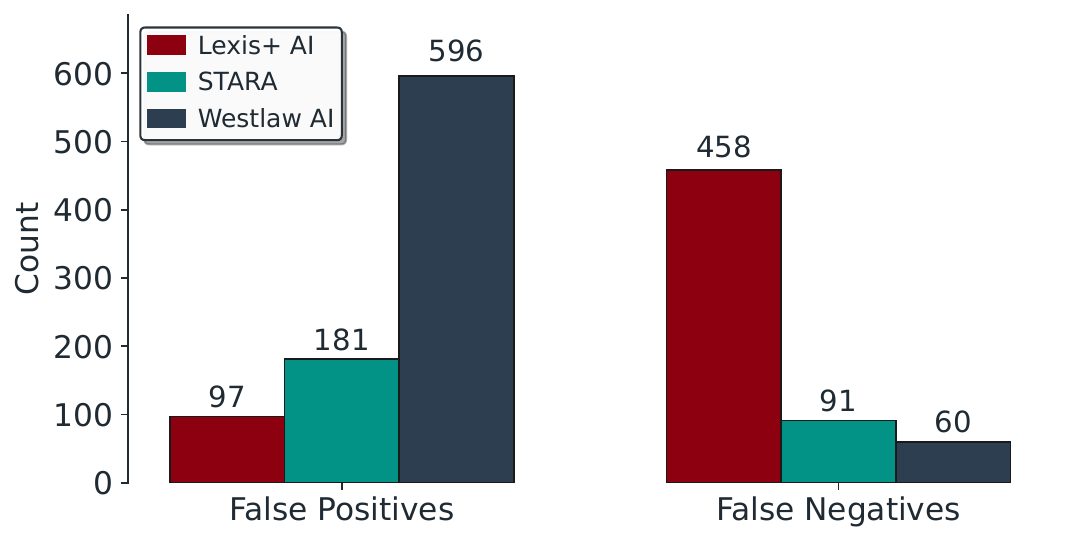}
        \caption{Distribution of false positives and false negatives across Lexis+ AI, Westlaw AI, and STARA.}
        \label{fig:error_patterns}
    \end{minipage}
\end{figure*}

We evaluate system performance in Table~\ref{tab:performance_comparison} using standard classification metrics: accuracy, precision, recall, and F1 score. We include a baseline representing a majority class classifier (\textit{i.e.}, answering affirmatively for all questions), as well as the best performing RAG model tested by \citet{haririAIStatutorySimplification2025}.
Our evaluation reveals significant performance differences across legal AI systems. STARA achieves 83\% accuracy, outperforming Westlaw AI and Lexis+ AI by 25 and 19 percentage points respectively. STARA maintains balanced precision and recall, demonstrating consistent accuracy across the full set of questions. We describe the corrected performance of STARA in Section~\ref{sec:results-fp}. Figure~\ref{fig:error_patterns} visualizes the distribution of false positives and false negatives across all three systems. While STARA produced approximately twice the false positives of Lexis+ AI (181 vs. 97), Westlaw AI generated over three times STARA's count with 596 total false positives.

\subsection{Comparative System Output Analysis}
\label{sec:results-error}

We examine error patterns across all three systems through detailed analysis of three representative question categories: self-employment assistance programs, state authority to deduct food stamp benefit overissuances, and alternative base period availability. We conclude this section with a summary of findings.

\subsubsection{Self-Employment Assistance}
\label{sec:results-error-sea}

\begin{figure*}[!htbp]
\centering
\includegraphics[width=\textwidth]{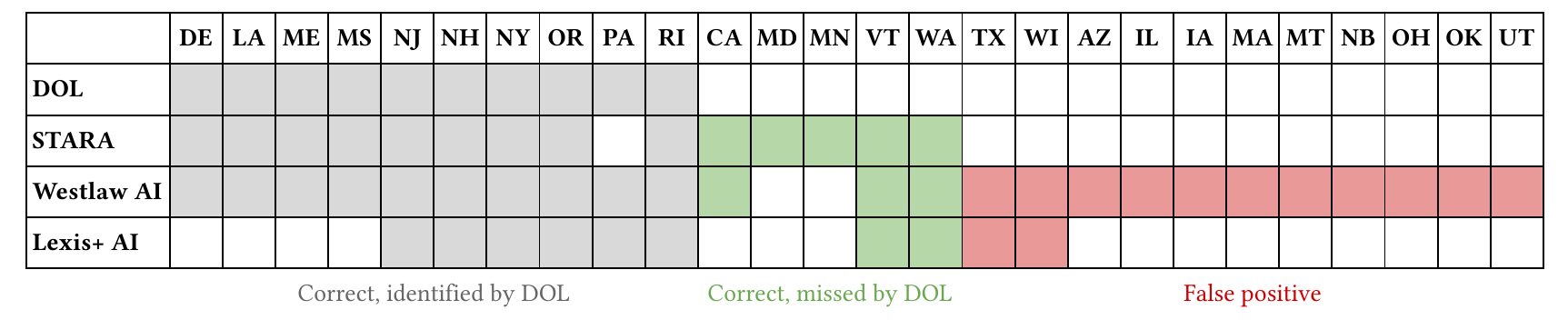} 
\caption{Comparative performance on identifying states with self-employment assistance programs, including both active programs and authorizing legislation. STARA identified 14 total states, 9 from Department of Labor (DOL) compilation plus 5 additional. Westlaw AI showed higher recall but numerous false positives. Lexis+ AI identified 8 states with high precision but low recall.}
\label{fig:example_question}
\end{figure*}

States with either active self-employment assistance programs, which allow unemployed individuals to start businesses while continuing to receive UI benefits, or authorizing legislation were evaluated following DOL methodology. As illustrated in Figure~\ref{fig:example_question}, STARA identified 9 of the 10 states in the DOL compilation and discovered five additional states: Maryland and Vermont had active programs, California and Washington had authorizing statutes without active programs, and Minnesota operates CLIMB \cite{minnesotalegislatureMinnesotaStatutes116L172024}, a functionally equivalent program that permits UI benefits during entrepreneurial training. STARA’s discoveries, each confirmed as a genuine statutory provision absent from the DOL report, led to extended verification of other benchmark questions to distinguish true system errors such as retrieval or reasoning errors from DOL omissions. Lexis+ AI identified only six of the original ten states and two of the additional states, demonstrating lower recall consistent with its performance across other questions. Westlaw AI identified more states but produced numerous false positives, underscoring the precision-recall tradeoff in multi-jurisdictional statutory analysis.

\subsubsection{SNAP Overissuance}
\label{sec:results-error-snap}


SNAP (Supplemental Nutrition Assistance Program, formerly food stamps) overissuances occur when recipients receive benefits exceeding their eligibility, creating debts that agencies seek to recover. The question asks whether states have statutory authority to deduct these SNAP debts from unemployment compensation payments, which requires explicit cross-program authorization linking the SNAP and UI systems. 

\begin{table}[ht]
\renewcommand{\arraystretch}{1.4}
\centering
\caption{System performance comparison by state.}
\label{tab:snap_comparison}
\begin{tabular}{|p{1.3cm}|p{2.9cm}|p{2.9cm}|}
\hline
\textbf{System} & \textbf{Alabama} (True) & \textbf{Alaska} (False) \\ \hline
STARA & 
\textit{True}. Cites statute (AL § 25-11-14). & 
\textit{False}. Correctly finds no statutory authority. \\ \hline
Westlaw AI & 
\textit{True}. Adds non-determinative references. & 
\textit{True} (FP). Points to unrelated sections. \\ \hline
Lexis+ AI & 
\textit{False}. Not in the list of states. & 
\textit{False}. Not in the list of states. \\ \hline
\end{tabular}
\end{table}

Table~\ref{tab:snap_comparison} compares system outputs for Alabama and Alaska, illustrating both accurate retrieval and common sources of error. STARA identifies the relevant statutory provisions with precise citations. Westlaw AI correctly identifies Alabama's authority but adds tangential provisions that do not alter the legal determination, increasing review effort (see Appendix~\ref{app:response-length}). In Alaska, Westlaw AI produces a false positive by citing SNAP recovery and child-support provisions that do not grant unemployment-insurance deduction authority. Lexis+ AI fails to capture valid authorities in both states.

Overall, STARA produced two apparent false positives on this question, as detailed in Table~\ref{tab:stara_fp_review}: West Virginia and Michigan. West Virginia was confirmed as correct and Michigan represented a reasoning error from STARA, where Mich. Comp. Laws § 421.11 authorizing information sharing with the U.S. Department of Agriculture was misclassified as deduction authority. Westlaw AI generated 21 apparent false positives, only one of which turned out to be correct. The 20 reasoning errors followed systematic patterns: misreading child support deduction statutes as SNAP authority, conflating UI overpayment recovery with cross-program offsets, and treating UI-to-UI interstate reciprocal arrangements as authorizing UI-to-SNAP deductions.

\begin{table}[!ht]
\renewcommand{\arraystretch}{1.25}
\setlength{\tabcolsep}{2.5pt} 
\centering
\caption{Verification of selected STARA apparent false positives on the SNAP overissuance deduction question. West Virginia shows a \textbf{Correct} finding missed in the DOL compilation, while Michigan reflects a \textbf{Reasoning Error} by STARA.}
\label{tab:stara_fp_review}
\begin{tabular}{|p{2cm}|p{6cm}|}
\hline
\textbf{State} & \textbf{Verification Outcome} \\
\hline
West Virginia &
Correct — W. Va. Code § 21A-6-17 authorizes the commissioner to deduct and withhold from unemployment compensation to recover food stamp overissuances. This authority was absent from the DOL compilation and confirmed as a correct identification. \\
\hline
Michigan &
Reasoning Error — Mich. Comp. Laws § 421.11 authorizes information sharing with the U.S. Department of Agriculture for the food stamp program but does not grant deduction authority; coordination was misclassified as substantive authority. \\
\hline
\end{tabular}
\end{table}

\subsubsection{Alternative Base Period}
\label{sec:results-error-abp}

The alternative base period question examines whether states provide alternative calculation methods for UI eligibility when claimants lack sufficient wages in the standard base period (typically the first four of the last five completed quarters). The DOL compilation documented 38 states with alternative base periods. STARA captured 35 (missing Arizona, Nevada, and Wisconsin) and identified one DOL omission (Missouri) plus one false positive (Wyoming). Lexis+ AI found 19 states, correctly identifying 15 from the DOL list with four false positives including Missouri (correctly identified by all systems), Alabama, Tennessee, and Texas. Westlaw AI identified 49 states, capturing 37 of the DOL's 38 (missing South Dakota) with 12 false positives. See Appendix~\ref{app:alternative-base-period} for detailed analysis.

\subsubsection{Summary}
\label{sec:results-error-summary}

Table~\ref{tab:3_reps} summarizes verification results across these three questions, showing the proportion of apparent false positives that were confirmed as valid state authorities missed by the DOL compilation, with STARA far exceeding the two commercial platforms in actual validity rate. Appendices~\ref{app:multi-quarter} and~\ref{app:voluntary-contributions} document additional false positive analysis and reasoning error patterns for multi-quarter calculations and voluntary contributions.

\begin{table}[!htbp]
\renewcommand{\arraystretch}{1.25}
\setlength{\tabcolsep}{2.5pt} 
\centering
\caption{Out of apparent false positives across three representative questions, number and \% confirmed as legitimate DOL omissions.}
\label{tab:3_reps}
\begin{tabular}{|l|c|c|c|}
\hline
\textbf{Question} & \textbf{STARA} & \textbf{Lexis+ AI} & \textbf{Westlaw AI} \\
\hline
SEA authorization & 5/5 (100\%) & 2/4 (50\%) & 3/14 (21\%) \\
\hline
SNAP offset authority & 1/2 (50\%) & 0/0 (—) & 1/21 (5\%) \\
\hline
Alternative base period & 1/2 (50\%) & 1/4 (25\%) & 1/12 (8\%) \\
\hline
\textbf{Total} & \textbf{7/9 (77\%)} & \textbf{3/8 (37\%)} & \textbf{5/47 (10\%)} \\
\hline
\end{tabular}
\end{table}

\begin{figure}[!ht]
\centering
\includegraphics[width=0.5\textwidth]{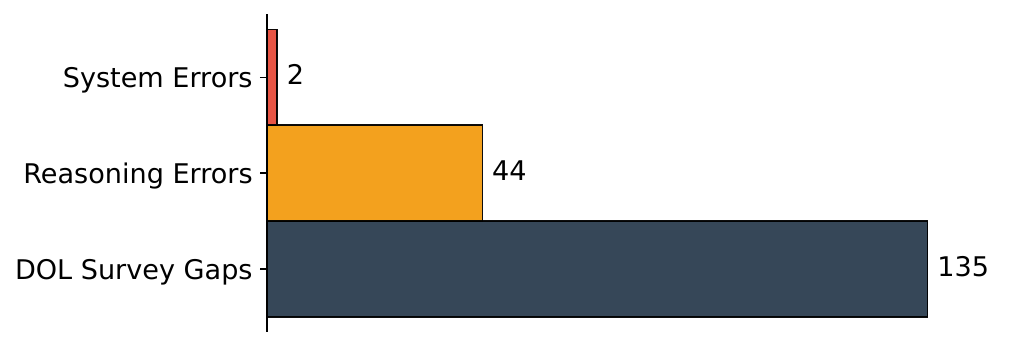}
\caption{STARA false positives by error type. DOL Survey Gaps represent legitimate omissions from the expert compilation, Reasoning Errors indicate misclassification of legal provisions, and System Errors reflect technical mistakes in cross-state citation processing.}
\label{fig:stara-false-positives}
\end{figure}

Considering the overall performance comparison in Section~\ref{sec:results-overall} and the selective examination in Section~\ref{sec:results-error}, we choose to focus a more comprehensive error analysis on STARA. We do so for three reasons. First, as illustrated in Figure~\ref{fig:error_patterns}, across all benchmark questions, STARA produced 181 false positives compared to Westlaw AI's 596 and Lexis+ AI's 97, making comprehensive manual review tractable for STARA while prohibitive for Westlaw AI. Second, Lexis+ AI produced the fewest false positives overall (97) and showed substantial overlap with STARA's false positives. Third, as detailed in Table~\ref{tab:3_reps}, Westlaw AI's 10\% accuracy (5 correct out of 47 apparent false positives) compared to STARA's 77\% (7 correct out of 9) suggests that Westlaw AI's errors predominantly reflect systematic reasoning failures rather than actual DOL compilation gaps. We note that this error analysis can be time consuming, as it requires substantive review of and engagement with complex UI provisions. 

We proceed with comprehensive validation of STARA's apparent false positives and false negatives, followed by additional analysis of commercial platform limitations.

\subsection{STARA Output Validation}
\label{sec:results-stara}

\subsubsection{Analysis of STARA's False Positives}
\label{sec:results-fp}

135 out of STARA's 181 apparent false positives actually reflect correct identifications of statutory provisions that were missed in the DOL compilation. As shown in Figure \ref{fig:stara-false-positives}, manual verification of every flagged case confirms that the vast majority stem from gaps in the expert compilation rather than reasoning or system errors. 
The corrected performance metrics, which reflect these findings, are included as the last row of Table \ref{tab:performance_comparison}.
Many of the 44 entries tagged as reasoning errors arise from how the benchmark treats older but still codified provisions. A statute can remain in the code with historical effect only, and the prompt does not specify whether such expired provisions should count as True. Table \ref{tab:historical_scope_examples} shows two examples from Michigan and Indiana where this ambiguity leads to answers marked as reasoning mistakes. Finally, system errors in Figure \ref{fig:stara-false-positives} comprise two cases where STARA cited provisions from the wrong state while answering a state-specific question. 

\begin{figure*}[!htbp]
\centering
\includegraphics[width=.9\textwidth]{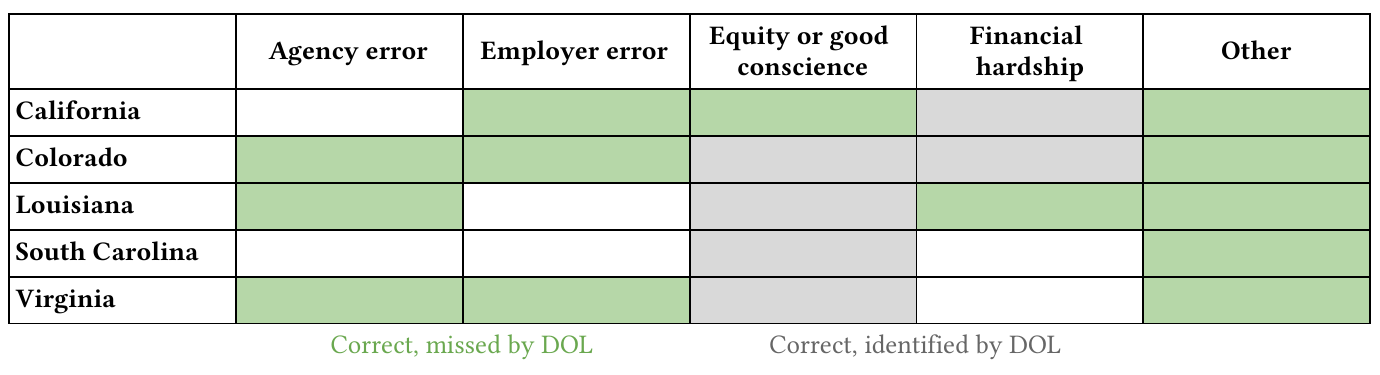}
\caption{Overpayment waiver grounds for five illustrative states.}
\label{fig:waiver_matrix_examples}
\end{figure*}

\begin{table}[H]
\centering
\renewcommand{\arraystretch}{1.3}
\caption{Two illustrative cases where scope limits drive label decisions, for a question on how states calculate weekly benefit amounts.}
\label{tab:historical_scope_examples}
\begin{tabular}{|p{0.95\columnwidth}|}
\hline
\textbf{Michigan historical weekly wage method} \\
\hline
\textbf{STARA answer:} \textit{``Michigan uses a weekly wage method. Section 421.27 Payment of Benefits states that the weekly benefit rate for an individual, with respect to benefit years beginning before October 1, 2000, is 67\% of the individual's average after tax weekly wage.''} \\[6pt]
\textbf{Clarification:} Applies only to benefit years beginning before Oct 1, 2000; retained in the code as a historical rule. \\
\hline
\textbf{Indiana historical high quarter method} \\
\hline
\textbf{STARA answer:} \textit{``Indiana historically used a high quarter method for claims filed before July 2012. 22 4 12 2 Rates; Prior Weekly Wage Computation describes a calculation based on percentages of wage credits in the calendar quarter during the individual's base period in which the wage credits were highest.''} \\[6pt]
\textbf{Clarification:} Applies only to claims filed before July 2012; retained in the statute for historical reference. \\
\hline
\end{tabular}
\end{table}


\subsubsection{Analysis of DOL Compilation Gaps}
\label{sec:results-dol}

The 135 corrected omissions by DOL are concentrated in certain question types rather than spread evenly across the benchmark. The largest share involve overpayment–waiver questions, as illustrated in Figure~\ref{fig:waiver_matrix_examples}. Whether or not repayment can be waived for agency or employer error, financial hardship, equity or good-conscience, or other grounds directly affects who must repay benefits and under what conditions, and erroneous statutory surveys can misdirect legal research and agency oversight. 


Beyond waivers, additional confirmed gaps appeared in questions on self-employment assistance, alcohol or drug disqualification, extended base period availability, part-time work search, and related topics, as summarized in Table~\ref{tab:missed_provisions}. Each provision category is a clear statutory concept with varying articulation across states, which complicates systematic identification.  Several factors help explain these omissions by DOL. Variation in statutory drafting can hide a common concept when key terms differ across states. Relevant language may be placed in definitions or cross-references that are easy to overlook when building state-by-state tables. Keyword search can fail when phrasing is atypical, and the scale of reviewing fifty separate codes makes it difficult to capture every scattered provision even with careful legal analysis. These factors match the patterns seen in Figure~\ref{fig:waiver_matrix_examples} and in the other categories aforementioned.

\begin{table*}[!ht]
\centering
\caption{Other categories of statutory provisions with state findings omitted by DOL.}
\small
\begin{tabular}{|p{4cm}|p{7.5cm}|p{3.4cm}|}
\hline
\textbf{Provision Category} & \textbf{Description} & \textbf{Additional States Identified} \\
\hline
Part-Time Work Search & Part-time work search satisfies availability or search requirements & CA, MI, MT\\
\hline
Self-Employment Assistance & Programs allowing continued UI benefits while pursuing self-employment & CA, MD, MN, VT, WA\\
\hline
Extended Base Period Availability & States allowing extended base period for eligibility calculation & NH, NJ, NM, NY, VT, WA\\
\hline
Loan and Interest Repayment Taxes & State imposes special loan or interest repayment taxes for UI & IL, IN, KS, MA, MT, NC, OH, RI, SD, UT  \\
\hline
Alcohol or Drug Disqualification & Disqualification or case-specific procedures for alcohol or drug-related terminations & CO, ID, IN, IA, ME, MA, MN, MS, NC, TX, VT, WI\\
\hline
Retirement Payments Exclusion & Excludes retirement payments from affecting base-period work if not affected by base-period employment & AR, CO, DE, IL, LA, MD, MN, MS, NM, OR, SC, SD, OH, VA, VT, WY \\
\hline
\end{tabular}
\label{tab:missed_provisions}
\end{table*}

\begin{figure*}[t]
\centering
\includegraphics[width=1\textwidth]{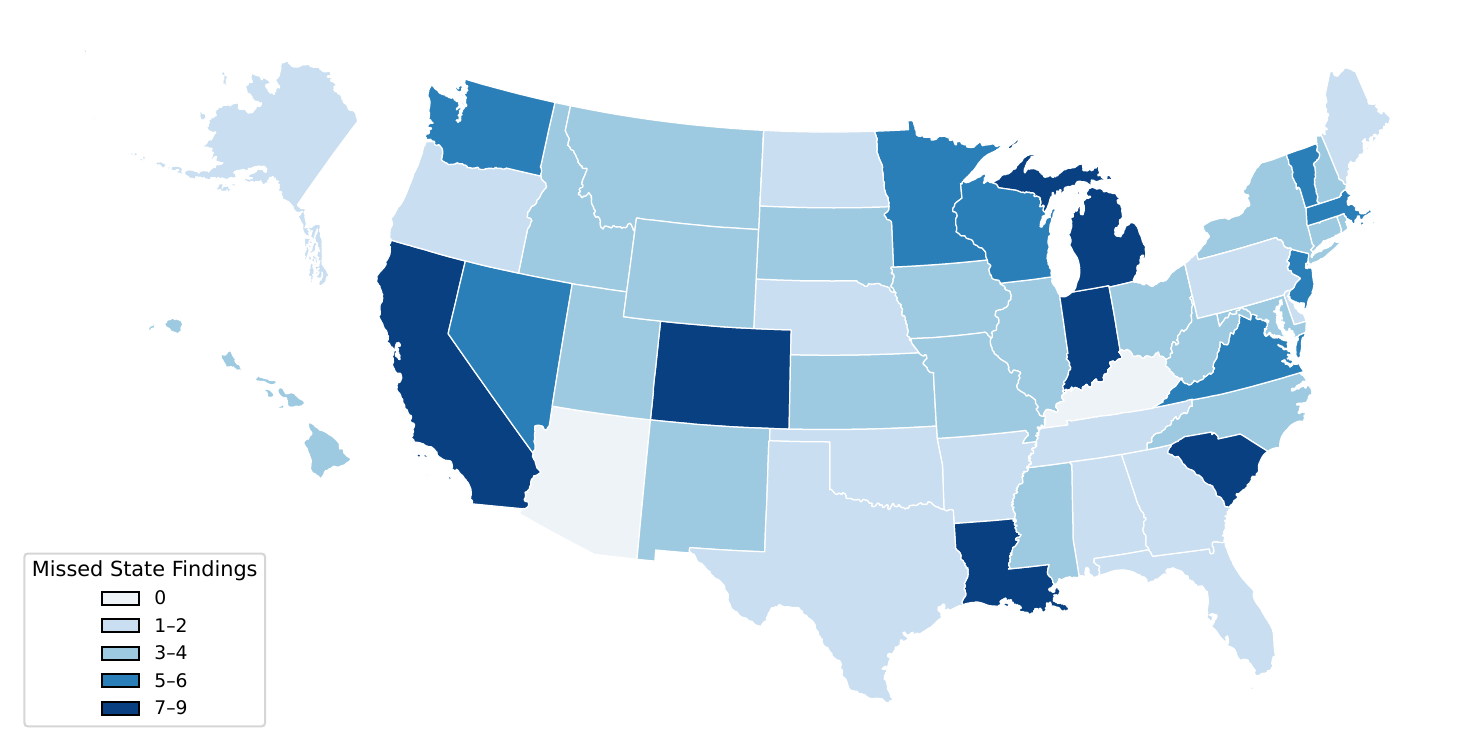}
\caption{Geographic distribution of state findings missed in the Department of Labor compilation.}
\label{fig:missed_provisions_map}
\end{figure*}

Figure~\ref{fig:missed_provisions_map} reveals significant geospatial variation in DOL compilation gaps, ranging from zero to nine missing state findings across the forty question categories evaluated. Arizona and Kentucky were the only states where STARA found no additional provisions beyond those reported by DOL. Several other states, including Alaska, Arkansas, Georgia, and North Dakota, had only one missing provision each, suggesting relatively comprehensive coverage in the original compilation. In contrast, 
Michigan's nine missing provisions are particularly alarming, representing approximately 29\% of the 31 total questions evaluated for that state in the LaborBench dataset and indicating substantial gaps in the original compilation's coverage of Michigan's UI statutory framework.

\subsubsection{Analysis of STARA's False Negatives}
\label{sec:results-fn}

Compared to the apparent false positives, where the system provides statutory reasoning and a source that can be checked against the DOL compilation, false negatives represent a distinct challenge as STARA did not retrieve or identify any relevant provisions for a given question in a state. Across the benchmark it produced 69 such misses, and 24 of them came from only two questions: \textit{(1)} whether part-time work search is acceptable in the state, and \textit{(2)} whether the state expands the coverage provisions for nonprofit organizations beyond federal requirements. 
This concentration suggests that the problem lies primarily in retrieval rather than reasoning. Federal law requires coverage for services performed for religious, charitable, or educational nonprofit organizations only when such organizations employ four or more workers over twenty weeks in a year, and states that broaden this requirement often use highly varied statutory language. Because the initial evaluation applied RegEx filters to narrow the search space for computation, relevant provisions using different phrasing were sometimes excluded before classification. Given that these two questions alone account for roughly one-third of all missed findings, false negatives appear to reflect the limits of filtering and linguistic variation rather than a broader weakness in STARA’s reasoning.

Another source of false negatives relates to the scope of legal materials included in the evaluation. Some states establish certain  provisions like part-time work eligibility through regulations or administrative interpretations rather than statutory law. The DOL compilation captures these non-statutory authorities and marks them accordingly (using notation like ``R'' for regulation or ``I'' for interpretation in their source documentation). However, since STARA searches only statutory text for this benchmark, states that recognize part-time work eligibility solely through regulation or administrative policy would not be detected. This limitation affects multiple jurisdictions including Utah, Oregon, and Nevada, where the operative rules exist in regulatory or policy documents rather than codified statutes.

\subsection{Analysis of Commercial Platform Limitations}
\label{sec:results-commercial}

Evaluation of commercial legal AI platforms reveals significant architectural constraints that limit their effectiveness. Such limitations manifest both in interface design and in the underlying processing capabilities of the systems.

Westlaw AI imposes a 300-character limit on query input, making it nearly impossible to specify the nuanced definitional criteria and contextual requirements necessary for accurate statutory analysis (see Appendix~\ref{app:input-examples}). The system also demonstrated a pronounced tendency toward false positive classifications, frequently returning affirmative results even when clear statutory evidence contradicted such findings. For example, the system indicated that states without self-employment assistance programs possessed such programs, and classified 41 states as having employer definitions that differ from federal standards. In actuality, these parrot the federal definition regarding monetary thresholds and minimum week requirements. These errors suggest that the system may identify topically related legal provisions without properly analyzing their substantive content or relationship to the specific question posed.

Lexis+ AI accepts queries of up to 5{,}000 characters, which allowed us to provide the full context from the benchmark dataset, so input length was not a limiting factor. However, the system severely favored speed over completeness, producing many false negatives and a recall of 0.29. 

\begin{table}[H]
\renewcommand{\arraystretch}{1.2}
\setlength{\tabcolsep}{3pt} 
\centering
\small
\caption{Lexis+ AI response consistency analysis on whether states exclude \textit{(1)} elected officials, \textit{(2)} members of the National Guard, or \textit{(3)} policymaking and advisory positions from UI provisions. Westlaw AI maintained 100\% consistency across all three question types.}
\label{tab:commercial_consistency}
\begin{tabular}{|p{2cm}|p{1.7cm}|p{4.5cm}|}
\hline
& \textbf{Consistency Rate} & \textbf{Response Changes} \\
\hline
Elected Officials & 87.5\% & True to False: Nevada \\
\hline
National Guard & 12.5\% & False to True: AK, AR, GA, NE, NV, WA \newline True to False: Montana \\
\hline
Policymaking & 50\% & False to True: AK, NV \newline True to False: FL, NE \\
\hline
Total & 50\% & Total: 12 response changes \\
\hline
\end{tabular}
\end{table}

Table \ref{tab:commercial_consistency} shows the results of our tests of internal consistency within Lexis+ AI when switching from querying all states to querying just three states at a time. Westlaw AI answered True to all 24 questions related to UI exclusions across both query modes, demonstrating a consistent tendency to return positive responses regardless of the query mode used, which resulted in seven incorrect responses. For Lexis+ AI, responses changed for 12 out of 24 questions, with the National Guard exclusion question showing particularly low consistency at 12.5\%. Lexis+ AI accuracy improved from 66\% in survey mode to 83\% in three-state mode for these questions, revealing that the platform can produce substantially different results depending on which query mode is selected.



Both commercial platforms advertise the ability to complete ``fifty state surveys in minutes,'' but the major quality issues we've identified through our evaluation may render such speed advantages moot if users must undertake substantial manual corrections. STARA required significantly more compute budget, taking on the order of days to complete the LaborBench benchmark evaluation,\footnote{STARA can process questions in batches, with our current evaluation using batches of five questions taking approximately 20 minutes to filter provisions and generate answers for one state. Completing a full fifty-state survey for a single question requires roughly 3.3 hours, or 16.6 hours for five questions. STARA currently supports running four tasks simultaneously, which reduces the time for a single question survey to approximately 50 minutes. Processing more than four tasks in parallel or larger question batches could further improve efficiency without affecting accuracy, as tasks run independently.} but still delivers substantial time and resource savings relative to the DOL's 6 months by a team of expert attorneys, along with far more complete and precise results. 

\section{Discussion}
\label{sec:discussion}

\subsection{Effective Multi-Jurisdictional Statutory Surveys}

The stark performance differences between STARA and commercial platforms revealed by our study, such as accuracy gaps exceeding 24 percentage points, as well as the discovery that 75\% of STARA's apparent false positives actually represented valid statutory provisions absent from expert compilation, underscore the complexity of statutory surveys and the critical role of system design. We distill the insights from our evaluation into concrete design principles for both commercial platforms and research tools seeking to provide accurate multi-jurisdictional legal analysis.

\subsubsection{Precise question specification and contextual clarity.} Multi-jurisdictional surveys require explicit parameters defining temporal scope, handling of expired provisions, and treatment of exceptions. 
As exemplified in Table~\ref{tab:historical_scope_examples}, our benchmark revealed numerous cases where provisions applicable before specific dates remained in statute books, creating ambiguity about their classification. 
Well-designed survey prompts should state the evaluation window, specify whether historical provisions still codified count as positive findings, and clarify whether enumerated exception conditions control outcomes. Without these specifications, identical statutory texts may yield contradictory interpretations across systems. Clarifying these elements at the question level yields determinations that better align with how practitioners read and apply the statutes.

\subsubsection{Domain expertise in statutory interpretation.} Effective surveys require understanding how legal concepts manifest across jurisdictions. The definition of ``employer'' illustrates this challenge: states may maintain identical monetary thresholds while varying exemptions for agricultural workers, domestic employees, or nonprofit organizations. Systems must recognize whether such variations constitute meaningful differences for the survey's purpose. Questions about ``differences from federal definitions'' require specifying whether any deviation counts or only substantive changes to core requirements matter.

\subsubsection{Strategic corpus selection and retrieval scope.} Running searches across entire state codes rather than targeted statutory sections can introduce reasoning errors or computational overhead. Both Lexis+ AI and Westlaw AI frequently cited provisions containing similar keywords but addressing unrelated legal domains, such as employment discrimination statutes when searching for UI or workers' compensation provisions when querying benefit calculations. Restricting searches to relevant code titles or chapters may improve both efficiency and accuracy in these cases. By contrast, when STARA searched complete codes versus UI-specific provisions, it consistently retrieved correct UI statutes, demonstrating that semantic search effectively identifies relevant material even within massive corpora. However, computational overhead can severely limit exhaustive searches, so STARA involves the use of carefully designed RegEx filters which tradeoff between cost and potential omissions of relevant provisions (see Appendix~\ref{app:regex}).

\subsubsection{Transparent retrieval and citation practices.} Systems must provide traceable paths from questions to statutory text, enabling verification of both positive and negative findings. STARA's explicit citations allowed complete review of apparent errors, revealing that 75\% represented genuine provisions absent from expert compilation. Commercial platforms' lengthy outputs mixing relevant and tangential sources complicate validation and increase practitioner review burden (see Appendix~\ref{app:response-length}).

\subsubsection{Recognition of non-statutory authorities.} Many state implementations rely on regulations, administrative interpretations, or policy guidance rather than statutory text. 
As we observed in our analysis of false negatives, benchmarks whose ground truth incorporates regulations and administrative guidance alongside statutes will systematically disadvantage systems that search only statutory text. The measured ``errors'' in such cases reflect differences in the legal materials searched rather than deficiencies in legal reasoning. Evaluations should explicitly document whether their ground truth includes non-statutory sources, as this distinction fundamentally affects how system performance should be interpreted. A key missing element from DOL's statutory compilation are citations to the underlying legal authority, which our work begins to fill. 

\subsection{Limitations}
\label{sec:limitations}

Several limitations constrain the scope and interpretation of our findings while charting useful directions for future work. This evaluation benchmarks AI tools on multi-jurisdictional statutory surveys in UI law, a single legal domain with distinctive concepts and drafting patterns. While this area provides substantial complexity, performance here may not predict how the same systems would perform when conducting surveys in other areas of law where statutory structure, terminology, and jurisdictional variation differ. The benchmark assigns binary True or False labels for scoring, providing consistent metrics but inevitably simplifying the reasoning involved in multi-jurisdictional statutory analysis. Even with our review of supporting citations and explanations, this framework reduces complex legal interpretation to a single outcome. Future research could build on this work by introducing graded evaluations or multi-stage questions that measure not only correctness but also the quality of statutory reasoning and the integration of multiple provisions, offering a closer representation of real legal research.

Our verification process had inherent scope limitations. While we systematically reviewed all apparent false positives from STARA, resource constraints prevented comprehensive examination of the substantially larger volume of apparent false positives generated by Westlaw AI and Lexis+ AI. The commercial platforms produced outputs that were often lengthy and required extensive manual review to verify citations, making exhaustive validation impractical. Although our sampling of Westlaw AI results confirmed frequent misgrounded citations, some unexamined flags could potentially represent correct provisions absent from the DOL compilation. Additionally, our review focused on discrepancies between system outputs and the DOL compilation, meaning that provisions classified as absent by both STARA and the DOL were not systematically rechecked. This leaves open the possibility that some states contain relevant statutory provisions that all sources overlooked. Our study underscores that benchmarking is inherently difficult in real-world domains where human experts also make errors and omissions, as corroborated by other efforts to benchmark STARA against lists of federal crimes, congressionally mandated reports, and city commissions \cite{suraniWhatLawSystem2025a}.  

Evaluation of commercial AI systems faced inherent limitations. Both Westlaw AI and Lexis+ AI function as black boxes where the underlying statutory databases and search algorithms remain undisclosed, making it impossible to determine whether errors stem from incomplete coverage or flawed retrieval methods. While both platforms provide reasoning with their answers, the lack of transparency about their statutory sources complicates error analysis. Additionally, Westlaw AI imposes character limits on queries that prevented testing certain complex questions, further constraining comprehensive benchmarking across all 1{,}647 questions. These commercial tools understandably protect proprietary methods, but this hinders rigorous assessment. Marketing claims of ``fifty-state surveys in minutes'' warrant scrutiny given our findings that both systems performed below baseline in F1 scores, suggesting that speed may compromise the careful statutory analysis required for reliable legal research.

\section{Conclusion}
\label{sec:conclusion}

This study evaluates statutory retrieval at scale using LaborBench, a benchmark rooted in real questions about unemployment law across all 50 states. STARA was tested against Westlaw AI and Lexis+ AI and delivered the strongest performance, reaching 83\% accuracy with balanced precision and recall. The comparison shows that a retrieval approach built around statutory structure can answer multi-jurisdictional questions with greater consistency than currently available commercial tools while highlighting the types of provisions that remain challenging to classify.

A surprising finding is that ``ground truth'' data compiled by DOL in fact omits a meaningful number of valid provisions. Verification of STARA’s apparent false positives against the codes themselves confirmed many as correct, which raises STARA’s measured accuracy to 92\% but illustrates the challenges of benchmarking in the real world, where even federal agency experts may miss critical statutory provisions. The additions include overpayment waivers, benefit calculation methods, self-employment assistance, part-time search rules, and other UI categories that matter directly for claimants and administrators. Being able to surface these provisions with specific citations provides a practical way to keep widely used reference sources complete and to support decisions that turn on the precise content of state law.

The results also set expectations for commercial platforms. Despite prominent claims about rapid multi-jurisdictional surveys, including a purported ``secret sauce that can't be matched'' \cite{noauthor_introducing_nodate}, the evaluated systems struggled to return accurate, comprehensive answers on many core questions and were constrained by input limits and weak handling of statutory context. A ``secret sauce'' is, naturally, not particularly conducive to transparency; perhaps companies should instead provide more rigorous documentation and evaluation results to support marketing claims. Independent benchmarking on real statutory data plays an essential accountability function before such tools are relied upon for high-stakes work. Future efforts should extend this style of evaluation to additional legal domains and pair retrieval improvements with clearer question specifications about temporal scope and exceptions, so that reported answers match how practitioners read and apply the law.

\begin{acks}
We thank Ananya Karthik, Emily Robitschek, Allison Casasola, Yasmine Mabene, and Dan Bateyko for helpful feedback and comments.
\end{acks}

\bibliographystyle{ACM-Reference-Format}
\bibliography{references.bib}

@misc{50StateSurveys2025,
  title = {50 {{State Surveys}} - {{Westlaw}}},
  year = {2025},
  urldate = {2025-09-28},
  howpublished = {https://legal.thomsonreuters.com/en/products/westlaw/50-state-surveys},
  langid = {american}
}

@misc{AIHallucinationCases2025,
  title = {{{AI Hallucination Cases Database}} -- {{Damien Charlotin}}},
  year = {2025},
  urldate = {2025-09-28},
  howpublished = {https://www.damiencharlotin.com/hallucinations/}
}

@misc{AILawTracker2025,
  title = {{{AI Law Tracker}}},
  year = {2025},
  urldate = {2025-09-28},
  howpublished = {https://www.polarislab.org/ai-law-tracker.html}
}

@misc{AIPowered50State2025,
  title = {{{AI-Powered}} 50 {{State Legal Surveys}}: {{Survey}} of {{Laws}} and {{Regulations Available Within Minutes}} from {{LexisNexis Prot{\'e}g{\'e}}}},
  shorttitle = {{{AI-Powered}} 50 {{State Legal Surveys}}},
  year = {2025},
  urldate = {2025-09-28},
  howpublished = {https://www.lexisnexis.com/community/insights/legal/b/product-features/posts/ai-powered-50-state-legal-surveys-survey-of-laws-and-regulations-available-within-minutes-from-lexisnexis-protege},
  langid = {english}
}

@misc{bainCuttingPolicyCruft2024,
  title = {Cutting {{Through}} ``{{Policy Cruft}}'' - {{Niskanen Center}}},
  author = {Bain, Ben and Tsang, Christine, Ben Bain},
  year = {2024},
  month = oct,
  journal = {Niskanen Center - Improving Policy, Advancing Moderation},
  urldate = {2025-09-30},
  langid = {english}
}

@misc{BrokenLawsUnprotected2009,
  title = {Broken {{Laws}}, {{Unprotected Workers}}: {{Violations}} of {{Employment}} and {{Labor Laws}} in {{America}}'s {{Cities}}},
  shorttitle = {Broken {{Laws}}, {{Unprotected Workers}}},
  year = {2009},
  month = sep,
  journal = {National Employment Law Project},
  urldate = {2025-09-28},
  langid = {american}
}

@inproceedings{zheng2025reasoning,
  title={A reasoning-focused legal retrieval benchmark},
  author={Zheng, Lucia and Guha, Neel and Arifov, Javokhir and Zhang, Sarah and Skreta, Michal and Manning, Christopher D and Henderson, Peter and Ho, Daniel E},
  booktitle={Proceedings of the 2025 Symposium on Computer Science and Law},
  pages={169--193},
  year={2025}
}

@misc{chalkidisLexGLUEBenchmarkDataset2022,
  title = {{{LexGLUE}}: {{A Benchmark Dataset}} for {{Legal Language Understanding}} in {{English}}},
  shorttitle = {{{LexGLUE}}},
  author = {Chalkidis, Ilias and Jana, Abhik and Hartung, Dirk and Bommarito, Michael and Androutsopoulos, Ion and Katz, Daniel Martin and Aletras, Nikolaos},
  year = {2022},
  month = nov,
  number = {arXiv:2110.00976},
  eprint = {2110.00976},
  primaryclass = {cs},
  publisher = {arXiv},
  doi = {10.48550/arXiv.2110.00976},
  urldate = {2025-09-28},
  archiveprefix = {arXiv}
}

@article{dahlLargeLegalFictions2024b,
  title = {Large {{Legal Fictions}}: {{Profiling Legal Hallucinations}} in {{Large Language Models}}},
  shorttitle = {Large {{Legal Fictions}}},
  author = {Dahl, Matthew and Magesh, Varun and Suzgun, Mirac and Ho, Daniel E},
  year = {2024},
  month = jan,
  journal = {Journal of Legal Analysis},
  volume = {16},
  number = {1},
  pages = {64--93},
  issn = {2161-7201},
  doi = {10.1093/jla/laae003},
  urldate = {2025-09-28}
}

@misc{feiLawBenchBenchmarkingLegal2023,
  title = {{{LawBench}}: {{Benchmarking Legal Knowledge}} of {{Large Language Models}}},
  shorttitle = {{{LawBench}}},
  author = {Fei, Zhiwei and Shen, Xiaoyu and Zhu, Dawei and Zhou, Fengzhe and Han, Zhuo and Zhang, Songyang and Chen, Kai and Shen, Zongwen and Ge, Jidong},
  year = {2023},
  month = sep,
  number = {arXiv:2309.16289},
  eprint = {2309.16289},
  primaryclass = {cs},
  publisher = {arXiv},
  doi = {10.48550/arXiv.2309.16289},
  urldate = {2025-09-28},
  archiveprefix = {arXiv}
}

@article{guhaLegalBenchCollaborativelyBuilt2023b,
  title = {{{LegalBench}}: {{A Collaboratively Built Benchmark}} for {{Measuring Legal Reasoning}} in {{Large Language Models}}},
  shorttitle = {{{LegalBench}}},
  author = {Guha, Neel and Nyarko, Julian and Ho, Daniel and R{\'e}, Christopher and Chilton, Adam and K, Aditya and {Chohlas-Wood}, Alex and Peters, Austin and Waldon, Brandon and Rockmore, Daniel and Zambrano, Diego and Talisman, Dmitry and Hoque, Enam and Surani, Faiz and Fagan, Frank and Sarfaty, Galit and Dickinson, Gregory and Porat, Haggai and Hegland, Jason and Wu, Jessica and Nudell, Joe and Niklaus, Joel and Nay, John and Choi, Jonathan and Tobia, Kevin and Hagan, Margaret and Ma, Megan and Livermore, Michael and {Rasumov-Rahe}, Nikon and Holzenberger, Nils and Kolt, Noam and Henderson, Peter and Rehaag, Sean and Goel, Sharad and Gao, Shang and Williams, Spencer and Gandhi, Sunny and Zur, Tom and Iyer, Varun and Li, Zehua},
  year = {2023},
  month = dec,
  journal = {Advances in Neural Information Processing Systems},
  volume = {36},
  pages = {44123--44279},
  urldate = {2025-09-28},
  langid = {english}
}

@article{guhaStateStatutesProject2024,
  title = {The {{State Statutes Project Special Issue}}: {{Public Law}} in the {{States}}},
  shorttitle = {The {{State Statutes Project Special Issue}}},
  author = {Guha, Neel and Zambrano, Diego A.},
  year = {2024},
  journal = {Wisconsin Law Review},
  volume = {2024},
  number = {5},
  pages = {1615--1636},
  urldate = {2025-09-30},
  langid = {english}
}

@article{hamillCertainDeathFiftyState2007,
  title = {As {{Certain}} as {{Death}}: {{A Fifty-State Survey}} of {{State}} and {{Local Tax Laws}}},
  shorttitle = {As {{Certain}} as {{Death}}},
  author = {Hamill, Susan},
  year = {2007},
  month = nov,
  journal = {Working Papers}
}

@misc{haririAIStatutorySimplification2025,
  title = {{{AI}} for {{Statutory Simplification}}: {{A Comprehensive State Legal Corpus}} and {{Labor Benchmark}}},
  shorttitle = {{{AI}} for {{Statutory Simplification}}},
  author = {Hariri, Emaan and Ho, Daniel E.},
  year = {2025},
  month = aug,
  number = {arXiv:2508.19365},
  eprint = {2508.19365},
  primaryclass = {cs},
  publisher = {arXiv},
  doi = {10.48550/arXiv.2508.19365},
  urldate = {2025-09-23},
  archiveprefix = {arXiv}
}

@misc{lockeCaseLawRetrieval2022,
  title = {Case Law Retrieval: Problems, Methods, Challenges and Evaluations in the Last 20 Years},
  shorttitle = {Case Law Retrieval},
  author = {Locke, Daniel and Zuccon, Guido},
  year = {2022},
  month = feb,
  number = {arXiv:2202.07209},
  eprint = {2202.07209},
  primaryclass = {cs},
  publisher = {arXiv},
  doi = {10.48550/arXiv.2202.07209},
  urldate = {2025-09-28},
  archiveprefix = {arXiv}
}

@article{mageshHallucinationFreeAssessingReliability2025b,
  title = {Hallucination-{{Free}}? {{Assessing}} the {{Reliability}} of {{Leading}} {{{\textsc{AI}}}} {{Legal Research Tools}}},
  shorttitle = {Hallucination-{{Free}}?},
  author = {Magesh, Varun and Surani, Faiz and Dahl, Matthew and Suzgun, Mirac and Manning, Christopher D. and Ho, Daniel E.},
  year = {2025},
  month = jun,
  journal = {Journal of Empirical Legal Studies},
  volume = {22},
  number = {2},
  pages = {216--242},
  issn = {1740-1453, 1740-1461},
  doi = {10.1111/jels.12413},
  urldate = {2025-09-28},
  langid = {english}
}

@misc{minnesotalegislatureMinnesotaStatutes116L172024,
  title = {Minnesota {{Statutes}}, {\S} {{116L}}.17, {{Subd}}. 11},
  author = {Minnesota Legislature},
  year = {2024},
  journal = {Revisor of Statutes, State of Minnesota},
  urldate = {2025-09-30},
  howpublished = {https://www.revisor.mn.gov/statutes/cite/116L.17\#stat.116L.17.11}
}

@article{morainStateLevelSupportTobacco2018,
  title = {State-{{Level Support}} for {{Tobacco}} 21 {{Laws}}: {{Results}} of a {{Five-State Survey}}},
  shorttitle = {State-{{Level Support}} for {{Tobacco}} 21 {{Laws}}},
  author = {Morain, Stephanie R. and Garson, Arthur and Raphael, Jean L.},
  year = {2018},
  journal = {Nicotine \& Tobacco Research},
  volume = {20},
  number = {11},
  eprint = {26772024},
  eprinttype = {jstor},
  pages = {1407--1411},
  publisher = {Oxford University Press},
  issn = {1462-2203},
  urldate = {2025-09-30}
}

@article{ouelletteCanAIHold2025,
  title = {Can {{AI Hold Office Hours}}?},
  author = {Ouellette, Lisa and Motomura, Amy and Reinecke, Jason and Masur, Jonathan},
  year = {2025},
  month = jan,
  journal = {Coase-Sandor Institute for Law \& Economics Research Paper Series}
}

@book{pahlkaRecodingAmericaWhy2023a,
  title = {Recoding {{America}}: {{Why Government Is Failing}} in the {{Digital Age}} and {{How We Can Do Better}}},
  shorttitle = {Recoding {{America}}},
  author = {Pahlka, Jennifer},
  year = {2023},
  month = jun,
  publisher = {{Henry Holt and Company}},
  isbn = {978-1-250-26676-7},
  langid = {english}
}

@misc{stettner14Workers2021,
  title = {1 in 4 {{Workers Relied}} on {{Unemployment Aid During}} the {{Pandemic}}},
  author = {Stettner, Andrew},
  year = {2021},
  month = mar,
  journal = {The Century Foundation},
  urldate = {2025-09-30},
  langid = {english}
}

@inproceedings{suraniWhatLawSystem2025a,
  title = {What {{Is}} the {{Law}}? {{A System}} for {{Statutory Research}} ({{STARA}}) with {{Large Language Models}}},
  shorttitle = {What {{Is}} the {{Law}}?},
  booktitle = {20th {{International Conference}} on {{Artificial Intelligence}} and {{Law}}},
  author = {Surani, Faiz and Gailmard, Lindsey A. and Casasola, Allison and Magesh, Varun and Robitschek, Emily J. and Ho, Daniel E.},
  year = {2025},
  urldate = {2025-09-28}
}

@article{w.hahnStateFederalRegulatory2000,
  title = {State and {{Federal Regulatory Reform}}: {{A Comparative Analysis}}},
  shorttitle = {State and {{Federal Regulatory Reform}}},
  author = {W. ~Hahn, Robert},
  year = {2000},
  month = jun,
  journal = {The Journal of Legal Studies},
  publisher = {The University of Chicago Press},
  issn = {0047-2530},
  doi = {10.1086/468098},
  urldate = {2025-09-30},
  copyright = {{\copyright} 2000 by The University of Chicago. All rights reserved.},
  langid = {english}
}

@misc{zeichnerIntroducingArtificialIntelligence2024,
  type = {Government {{Agency}}},
  title = {Introducing {{Artificial Intelligence Adjudicator Assistance}} ({{AIAA}}): {{A Research Initiative Exploring Ways}} to {{Streamline Work}} for {{Adjudicators}}},
  shorttitle = {Introducing {{Artificial Intelligence Adjudicator Assistance}} ({{AIAA}})},
  author = {Zeichner, Nikki and Perez,, Amy and Martin, Olivia and Surani, Faiz and Magesh, Varun and Rodolfa, Kit and Ho, Daniel E. and Bhaskar, Mihir},
  year = {2024},
  month = jan,
  journal = {US Department of Labor},
  urldate = {2024-07-10},
  langid = {english}
}

@misc{noauthor_introducing_nodate,
	title = {Introducing jurisdictional surveys on {Westlaw} {Edge}},
	url = {https://legal.thomsonreuters.com/en/insights/articles/jurisdictional-surveys-on-westlaw-edge},
	language = {en-US},
	urldate = {2025-10-01}
}

\newpage

\onecolumn

\appendix

\section{Reading Statutory Provisions}
\label{app:reading-provisions}

\begin{figure}[H]
\centering
\includegraphics[width=1\textwidth]{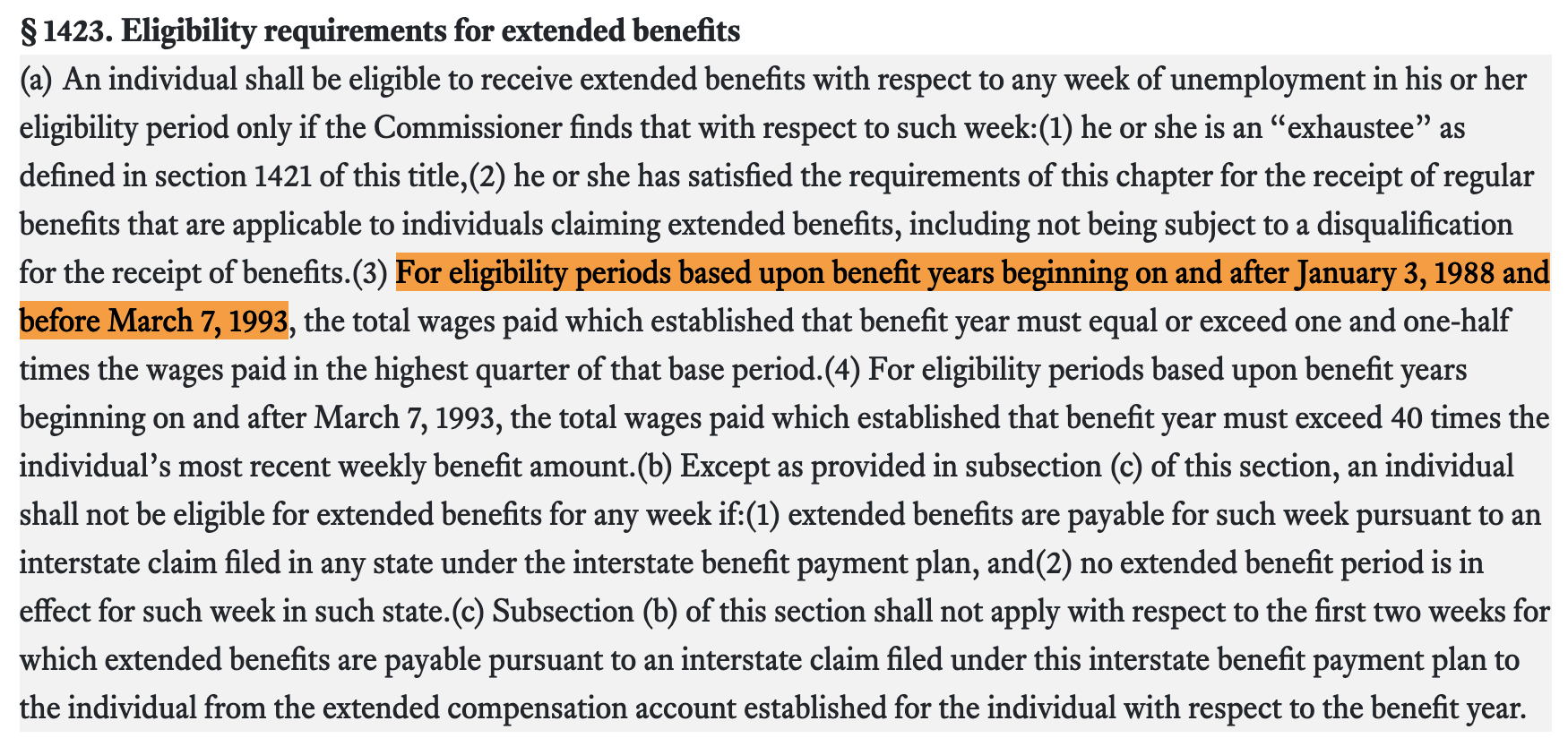}
\caption{Vermont Statutes Section 1423 showing superseded Extended Benefits eligibility requirement with temporal limitation.}
\label{fig:vermont_eb}
\end{figure}

Temporal qualifiers in statutory provisions require careful attention to ensure accuracy. Figure~\ref{fig:vermont_eb} shows Vermont Statutes Section 1423, which addresses Extended Benefits eligibility requirements. The question asks whether Vermont uses the alternative requirement of exceeding 1.5 times high-quarter wages to qualify for Extended Benefits. Subsection (a)(3) states that for eligibility periods based upon benefit years beginning on and after January 3, 1988 and before March 7, 1993, total wages must equal or exceed one and one-half times the wages paid in the highest quarter. However, subsection (a)(4) establishes that for eligibility periods based upon benefit years beginning on and after March 7, 1993, the requirement changed to total wages exceeding 40 times the individual's most recent weekly benefit amount. The 1.5 times high-quarter wages provision applied only from January 1988 through March 1993 and is no longer in effect. Vermont does not currently use this alternative requirement.

\begin{figure}[H]
\centering
\includegraphics[width=1\textwidth]{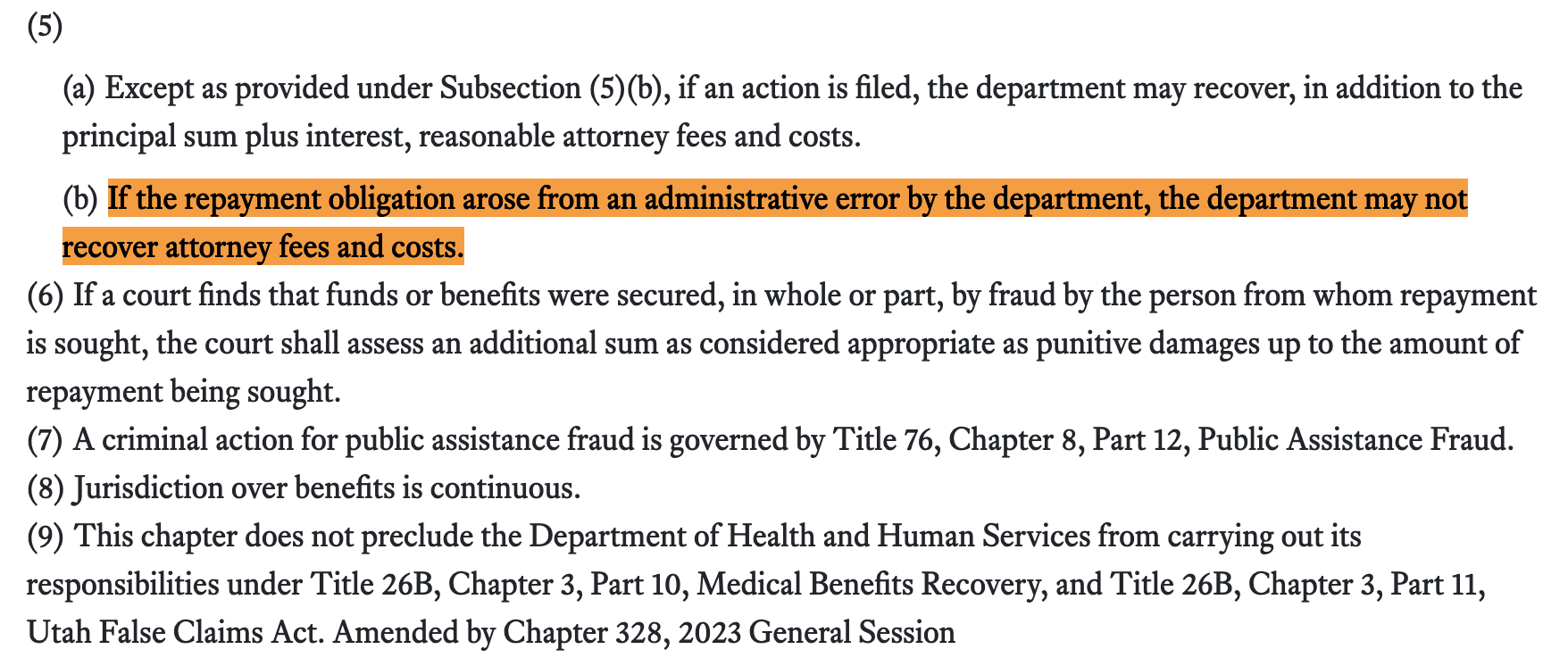}
\caption{Utah Code Section 35A-3-603 showing limited waiver of attorney fees and costs only, not overpayment principal.}
\label{fig:utah_partial}
\end{figure}

Some provisions create partial rather than complete waivers that require careful interpretation. Figure~\ref{fig:utah_partial} shows Utah Code Section 35A-3-603, which addresses civil liability for overpayments. The question asks whether Utah waives recovery of nonfraud overpayments if the overpayment was due to agency error. Subsection (5)(b) states that if the repayment obligation arose from an administrative error by the department, the department may not recover attorney fees and costs. However, this provision only waives the department's ability to collect fees and costs associated with recovery actions. The underlying overpayment itself remains collectible under subsection (1), which requires that a person who receives an overpayment shall, regardless of fault, return the overpayment or repay its value to the department. The administrative error provision does not waive the principal overpayment amount. Utah does not waive recovery of the overpayment itself when caused by agency error.

\begin{figure}[H]
\centering
\includegraphics[width=0.9\textwidth]{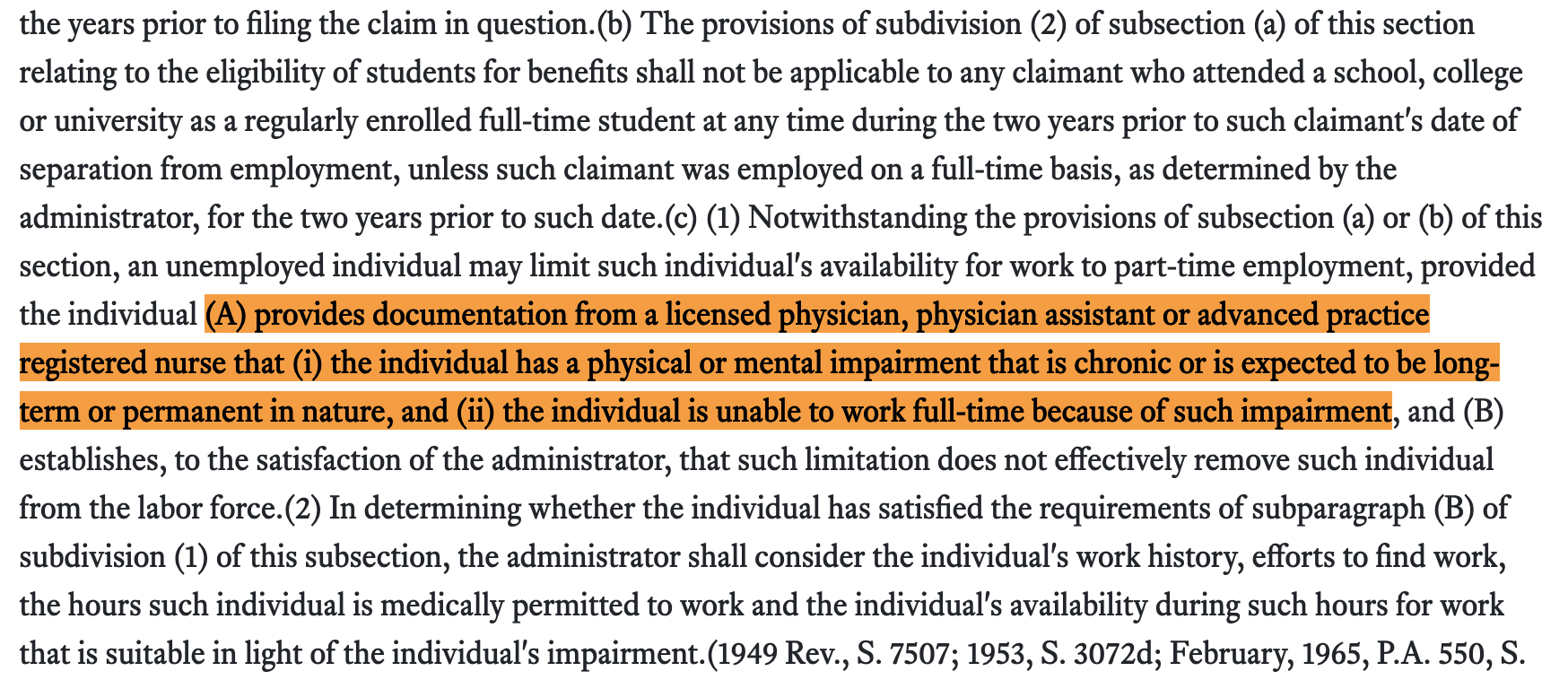}
\caption{Colorado Revised Statutes Section 8-73-108 showing medical exception for part-time work search as valid category for part-time eligibility.}
\label{fig:colorado_parttime}
\end{figure}

Medical exceptions to work search requirements represent a valid category for part-time work search eligibility. Figure~\ref{fig:colorado_parttime} shows Colorado Revised Statutes Section 8-73-108, which addresses work search requirements. The question asks whether part-time work search is acceptable in the state. Subsection (4)(o)(I) states that an individual shall not be disqualified from benefits for any week the individual is unable to work or seek full-time work if the individual provides medical documentation of a physical or mental impairment and the individual is able to work and is seeking part-time work. This provision allows part-time work search as a medical accommodation for individuals with documented impairments preventing full-time work. While not a general unrestricted allowance, medical-based part-time work search provisions are recognized as a distinct category for states allowing those seeking only part-time work to be eligible for unemployment compensation, alongside other categories such as claims based on part-time work history and unrestricted part-time work search eligibility.

\begin{figure}[H]
\centering
\includegraphics[width=0.9\textwidth]{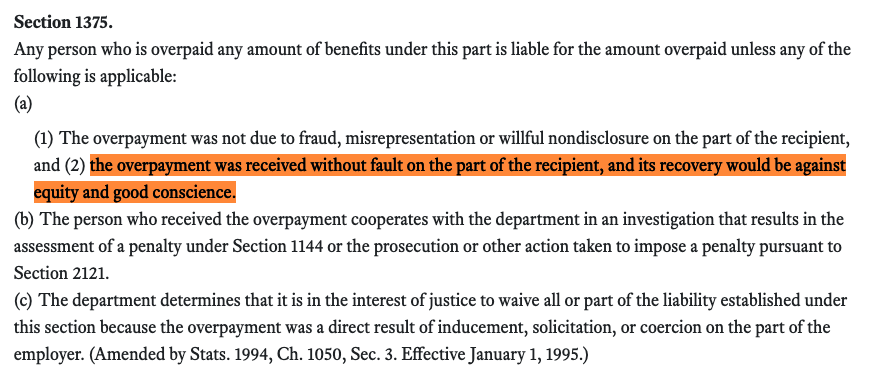}
\caption{California UI Code Section 1375 showing overpayment waiver provisions omitted from DOL compilation.}
\label{fig:california_waiver}
\end{figure}

Figure~\ref{fig:california_waiver} shows California UI Code Section 1375, which addresses overpayment liability and waiver conditions. The statutory text establishes that persons who receive overpaid benefits are liable for repayment unless specific conditions apply. Subsection (a) authorizes waivers when the overpayment was not due to fraud or willful nondisclosure, when overpayment was received without fault, and recovery would be against equity and good conscience. Subsection (c) provides waiver authority when overpayment resulted from employer inducement, solicitation, or coercion. The DOL compilation lists California as having only a financial hardship waiver, omitting these equity-based and employer-fault waiver grounds present in the statute.

\begin{figure}[H]
\centering
\includegraphics[width=0.9\textwidth]{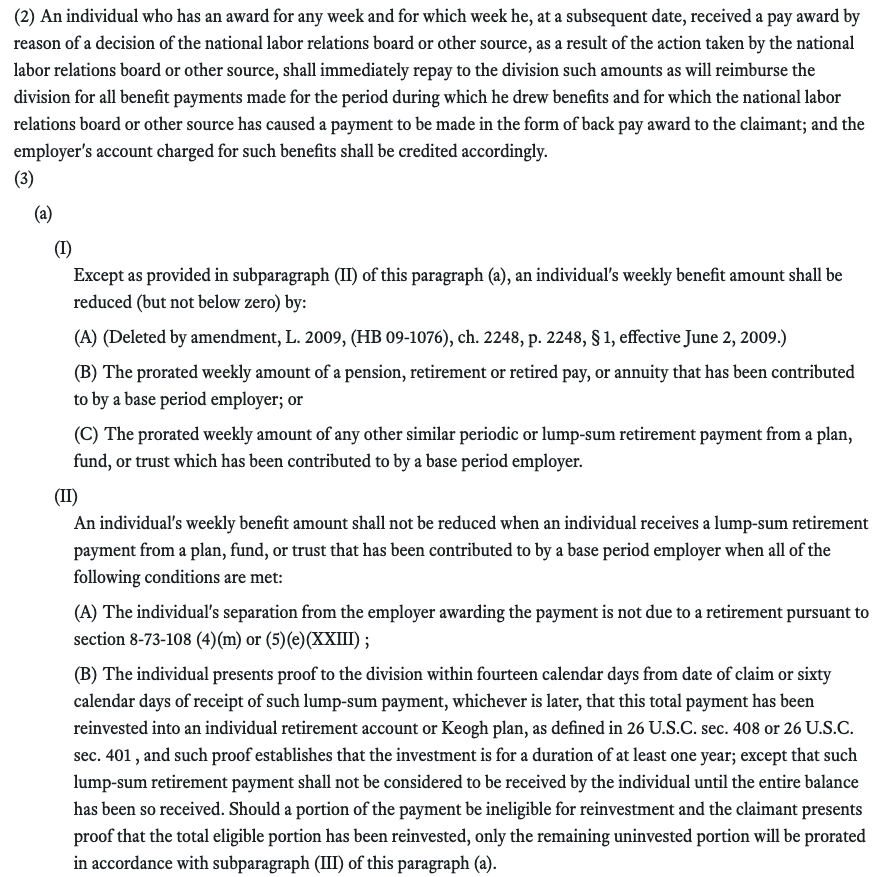}
\caption{Colorado Revised Statutes Section 8-73-110(3) showing retirement payment deduction provisions with base period employer qualifier.}
\label{fig:colorado_retirement}
\end{figure}

Some provisions require identifying specific qualifying conditions within lengthy statutory text. Figure~\ref{fig:colorado_retirement} shows Colorado Revised Statutes Section 8-73-110(3), which addresses retirement payment deductions from UI benefits. The question asks whether Colorado excludes retirement payments from affecting base period work if the payments are not influenced by base period work. The full provision addresses multiple payment scenarios and exceptions, but the key language appears in subparagraph (3)(a)(I): deductions apply to ``a pension, retirement or retired pay, or annuity that has been contributed to by a base period employer'' and ``any other similar periodic or lump-sum retirement payment from a plan, fund, or trust which has been contributed to by a base period employer.'' The repeated qualifier ``contributed to by a base period employer'' restricts deductions to retirement payments connected to base period employment. Retirement payments from non-base period sources fall outside this restriction and therefore do not affect base period work. Colorado excludes such payments.

\section{RegEx filters used for retrieval}
\label{app:regex}

STARA retrieval was restricted using RegEx filters applied to UI provisions, which typically appear in state labor code titles but sometimes under unemployment compensation or other organizational schemes that vary by state. 
Each filter uses two positive lookaheads. The first captures general unemployment insurance terminology. The second targets signals specific to each question batch. The benchmark applied eight total filters across forty question types, with each filter tailored to a batch of five related questions. Two example batches are provided below. Filters reduced computational requirements but created a tradeoff: those that use highly specific legal terminology or numeric thresholds can miss provisions drafted with variant phrasing, potentially contributing to the false negatives documented in Section~\ref{sec:results-fn}.

\subsection*{Batch 1: voluntary contributions, base period employer charging order, employer definition threshold, nonprofit coverage expansion, alcohol or drug provisions}

\begin{lstlisting}[breaklines=true,columns=fullflexible,keepspaces=true,showstringspaces=false,basicstyle=\ttfamily\large,linewidth=1.0\textwidth]
(?=.*(unemployment|UI|UC|employ|coverage|service|benefit|tax|contribution|experience.*rating|misconduct|discharge|disqualif))(?=.*(voluntary.*contribut|voluntary.*payment|additional.*contribut|improve.*rating|reduce.*rate|reserve.*ratio|benefit.*ratio|base.*period.*employer|charging.*employer|inverse.*chronolog|reverse.*order|last.*employer.*first|sequential.*charg|employer.*mean|employing.*unit.*that|one.*or.*more|four.*or.*more|\d+.*day|\d+.*week|\$\d+|1500|1000|20000|twenty.*week|10.*day|calendar.*quarter|nonprofit|religious|charitable|educational|501.*c.*3|church|exempt.*federal.*tax|FUTA.*3306|drug|alcohol|substance|controlled|intoxicat|test.*positive|refuse.*test|under.*influence))
\end{lstlisting}

\subsection*{Batch 2: reserve ratio experience rating, automatic benefit adjustments, overpayment waivers, retirement payment treatment, base period exclusions}

\begin{lstlisting}[breaklines=true,columns=fullflexible,keepspaces=true,showstringspaces=false,basicstyle=\ttfamily\large,linewidth=1.0\textwidth]
(?=.*(unemployment|UI|UC|benefit|employ.*security|experience.*rating|contribution|premium|tax.*rate|overpay|improper.*payment|retirement|pension|deduct|base.*period|weekly.*benefit|reserve|account|compensation.*fund))(?=.*(reserve.*ratio|reserve.*balance|contributions.*minus.*benefit|contributions.*paid.*less.*benefit|excess.*contribution|positive.*reserve|negative.*reserve|reserve.*surplus|reserve.*deficit|percent.*of.*excess|maximum.*benefit.*percent|maximum.*weekly.*benefit|average.*weekly.*wage|average.*weekly.*earning|automatic.*adjust|computed.*annually|ensuing.*twelve.*month|employment.*cost.*index|waiv|absolve|cancel|uncollectible|death|died|deceased|time.*limit|years.*following|de.*minimis|official.*advice|defeat.*purpose|retired|disabled|pro.*rata|proportional|employee.*contribution.*pension|employer.*funded|only.*employer.*paid|attributable.*to.*contribution|apply.*only.*if|does.*not.*apply|limited.*to.*base|only.*if.*base.*period))
\end{lstlisting}

\section{Example Prompts and Input Constraints}
\label{app:input-examples}

Commercial legal AI platforms impose varying input limitations that constrain their ability to process complex statutory questions. Lexis+ AI accepts queries up to 5,000 characters, allowing full context from the benchmark dataset, while Westlaw AI restricts input to 300 characters, requiring substantial compression of contextual information. The impact varies by question complexity: questions requiring minimal context, such as those about alcohol or drug provisions, face limited disadvantages, while questions requiring extensive definitional context face substantial constraints.

\begin{table}[H]
\renewcommand{\arraystretch}{1.25}
\setlength{\tabcolsep}{4pt} 
\centering
\caption{Prompt comparison for alcohol and drug provisions question.}
\label{tab:prompt_drugs}
\begin{tabular}{|l|p{12cm}|}
\hline
\textbf{System} & \textbf{Prompt} \\
\hline
STARA and Lexis+ AI &
\textit{Context:} A separation is considered involuntary in cases where there is a lack of work or reduction in force, or when an employer terminates the employment of an individual. In terminations from employment, the state looks to whether the individual engaged in misconduct to determine if the individual is eligible for UC. If a separation was not caused by any action or conduct of the individual, benefits would not be denied.

\textit{Question:} Does the state have provisions in their unemployment compensation law dealing specifically with alcohol and/or illegal drugs, and testing for alcohol or illegal drugs? \\
\hline
Westlaw AI &
States define gross misconduct as theft, assault, felonies, intoxication, safety violations affecting UC eligibility. Does the state include employers other than the last employer in determining disqualification for gross misconduct? Reply TRUE if yes to multiple employers, FALSE if no. \\
\hline
\end{tabular}
\end{table}

\begin{table}[H]
\renewcommand{\arraystretch}{1.25}
\setlength{\tabcolsep}{4pt} 
\centering
\caption{Prompt comparison for agricultural labor coverage question.}
\label{tab:prompt_agriculture}
\begin{tabular}{|l|p{12cm}|}
\hline
\textbf{System} & \textbf{Prompt} \\
\hline
STARA and Lexis+ AI &
\textit{Context:} The FUTA agricultural labor provisions apply to employing units who paid wages in cash of \$20,000 or more for agricultural labor in any calendar quarter in the current or preceding calendar year, or who employed 10 or more workers on at least one day in each of 20 different weeks in the current or immediately preceding calendar year. Under FUTA, agricultural labor is performed when workers raise or harvest agricultural or horticultural products on a farm, work in connection with the operation, management, conservation, improvement, or maintenance of a farm and its tools and equipment, handle, process, or package any agricultural or horticultural commodity if a farm produced over half of the commodity, do work related to cotton ginning or processing crude gum from a living tree, or do housework in a private home if it is on a farm operated for profit. The term ``farm'' includes stock, dairy, poultry, fruit, fur-bearing animals, and truck farms, as well as plantations, ranches, nurseries, ranges, greenhouses, or other similar structures used primarily for raising agricultural or horticultural commodities, and orchards. Agricultural labor does not include reselling activities that do not involve any substantial activity of raising agricultural or horticultural commodities. Most states have followed the FUTA provision and limited coverage to service performed on large farms. Any variation from these exact federal thresholds means the state has different agricultural coverage requirements.

\textit{Question:} Does the state have provisions for agricultural labor which differ from the FUTA 20 weeks/\$20,000 rule? \\
\hline
Westlaw AI &
Return TRUE or FALSE. FUTA baseline: \$20,000 in a quarter or 10 workers in 20 weeks. Does the state have provisions for agricultural labor which differ from the FUTA 20 weeks/\$20,000 rule? TRUE only if statute differs from the baseline in thresholds or weeks; if it matches, return FALSE. \\
\hline
\end{tabular}
\end{table}

Tables~\ref{tab:prompt_drugs} and~\ref{tab:prompt_agriculture} illustrate how input constraints affect prompt quality. The agricultural labor coverage question demonstrates how severe compression affects complex statutory analysis. STARA and Lexis+ AI received detailed context explaining FUTA provisions, the definition of agricultural labor, farm operations, coverage thresholds, and the specific variations that constitute differences from federal requirements. This context totaled over 1,500 characters and specified precisely what counts as a meaningful difference from federal standards. Westlaw AI's 300-character limit forced elimination of essential definitional context about what constitutes agricultural labor, what counts as a farm, which activities are excluded, and what types of variations matter. Without this framework, systems cannot reliably distinguish between meaningful statutory differences and superficial variations in phrasing. Questions requiring detailed legal specifications cannot be adequately conveyed within 300 characters, placing Westlaw AI at a fundamental disadvantage for complex statutory analysis.

\section{Response Length and Impact on Verification}
\label{app:response-length}

The length and structure of system responses significantly affect the effort required to verify outputs. For the SNAP overissuance deduction question, which asks whether states have statutory authority to deduct uncollected SNAP benefits from unemployment compensation, response formats varied substantially across platforms.

\begin{figure}[H]
\centering
\includegraphics[width=1\textwidth]{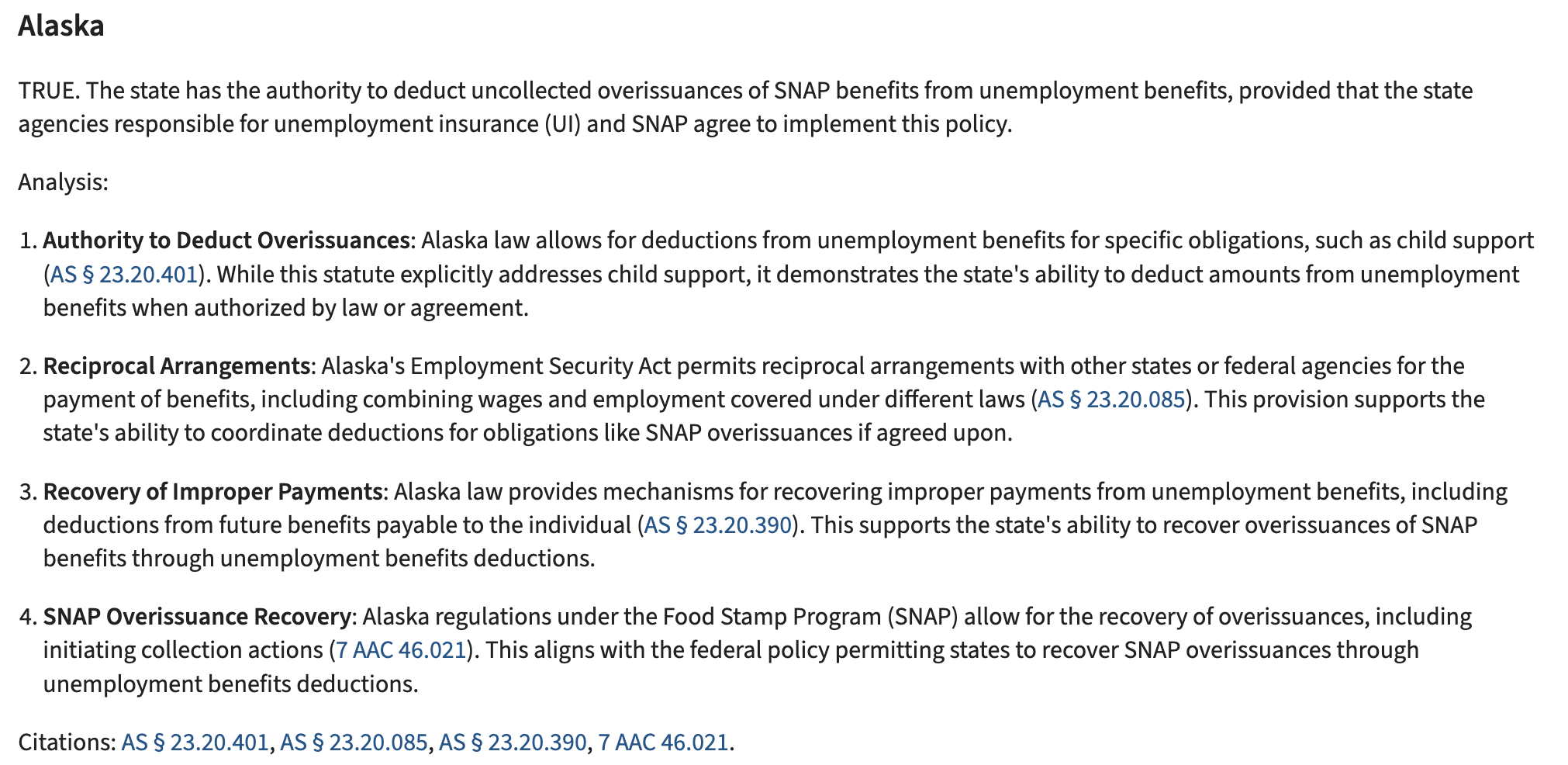}
\caption{Westlaw AI response for Alaska on SNAP overissuance deduction authority, showing a false positive with multiple unrelated statutory citations requiring extensive verification.}
\label{fig:westlaw_alaska_snap}
\end{figure}

Westlaw AI generated lengthy responses that cited multiple provisions, some tangentially related or unrelated to the specific question. Figure~\ref{fig:westlaw_alaska_snap} illustrates this verbose output format. For Alaska, where the correct answer is False, Westlaw AI produced a false positive citing four different statutory provisions. The response referenced AS § 23.20.401 (child support deductions), AS § 23.20.085 (reciprocal arrangements for interstate benefit payments), AS § 23.20.390 (recovery of improperly paid unemployment benefits), and 7 AAC 46.021 (SNAP program collection procedures). None of these provisions actually authorize deduction of SNAP overissuances from unemployment benefits. The child support statute addresses only Title IV-D support obligations. The reciprocal arrangements provision concerns UI-to-UI coordination between states. The improper payment recovery statute applies to UI overpayments within the same program. The SNAP regulation describes collection within the SNAP program itself. This verbose output requires extensive review to identify the reasoning error.

\begin{figure}[H]
\begin{center}
\includegraphics[width=0.85\textwidth]{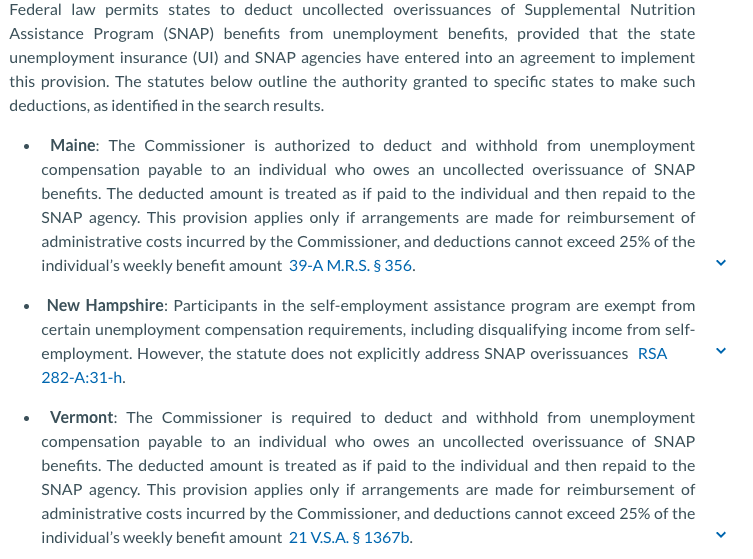}
\end{center}
\caption{Lexis+ AI response format for SNAP overissuance deduction question, showing concise outputs with specific statutory references and zero false positives but limited recall.}
\label{fig:lexis_snap}
\end{figure}

Lexis+ AI provided more concise responses than Westlaw AI and maintained significantly lower false positive rates. For this question, Lexis+ AI identified only three states as having SNAP deduction authority, all correct and included in the DOL compilation, producing zero false positives. Figure~\ref{fig:lexis_snap} illustrates this output format. However, Lexis+ AI achieved low false positive rates through conservative retrieval that resulted in numerous false negatives, missing many valid authorities captured by both STARA and the DOL compilation.

\begin{figure}[H]
\centering
\includegraphics[width=0.8\textwidth]{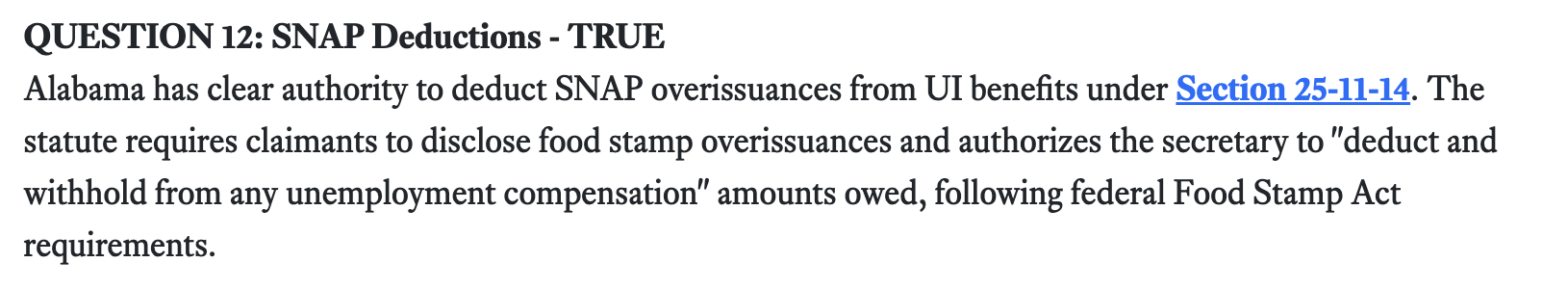}
\caption{STARA response for Alabama on SNAP overissuance deduction authority, showing concise citation to the authorizing statute.}
\label{fig:stara_alabama_snap}
\end{figure}

STARA provided concise responses with direct statutory citations. For Alabama, where the correct answer is True, STARA cited the specific authorizing provision (AL § 25-11-14) with minimal additional text, enabling rapid verification against the statute. Figure~\ref{fig:stara_alabama_snap} shows this streamlined output format.

The verification burden varies significantly with response length and citation quality. Westlaw AI's lengthy responses citing multiple tangentially related provisions require extended review to parse reasoning and identify errors. Lexis+ AI maintains verification efficiency through brevity but sacrifices completeness. STARA's focused citations enable quick validation against source statutes. These differences in output structure directly affect the practical utility of each system for legal research requiring verified answers.

\section{Alternative Base Period}
\label{app:alternative-base-period}

The alternative base period question identifies states that offer alternative calculation methods for determining UI eligibility when claimants have insufficient wages in the regular base period to qualify for benefits. The regular base period is typically the first four of the last five completed calendar quarters before filing a claim. The most common alternative is the last four completed quarters, though the DOL compilation documents several states offering multiple variations. Massachusetts allows the last three quarters plus any weeks of work in the quarter in which the claim is filed, with claimants also permitted to elect this alternative base period if it results in a 10\% or more increase in weekly benefit amount. New Jersey provides two alternatives: either the last four completed quarters or the last three completed quarters plus any weeks of work in the quarter in which the claim is filed. Vermont similarly offers two alternatives: the last four quarters or, if still ineligible, the last three quarters plus any weeks of work in the quarter in which the claim is filed. The alternative base period represents a different calculation method rather than an extension or modification of the time frame.

The DOL compilation identified 38 states with alternative base periods. STARA identified 35 of these states, missing Arizona, Nevada, and Wisconsin, and produced two apparent false positives in Missouri and Wyoming. Manual verification confirmed Missouri as a correct identification of a DOL omission, while Wyoming represented a reasoning error, yielding one actual DOL gap discovered by STARA. Lexis+ AI identified 19 states total, correctly capturing 15 of the 38 states in the DOL compilation and generating four apparent false positives: Missouri (correct, also identified by STARA and Westlaw AI), Alabama, Tennessee, and Texas. Westlaw AI identified 49 states as having alternative base periods, correctly capturing 37 of the 38 states in the DOL compilation while missing South Dakota, which the DOL report identifies as having an alternative base period of the last four completed quarters. Westlaw AI's 49 affirmative responses include 12 apparent false positives beyond the DOL compilation.

Tables~\ref{tab:stara_abp_fps} and~\ref{tab:lexis_abp_fps} present verification results for STARA's and Lexis+ AI's apparent false positives. Table~\ref{tab:westlaw_abp_errors} presents four representative reasoning errors from Westlaw AI's 12 apparent false positives, all of which proved to be actual errors following comprehensive verification.
\begin{table}[H]
\renewcommand{\arraystretch}{1.3}
\setlength{\tabcolsep}{3pt}
\centering
\small
\caption{Verification of STARA apparent false positives on alternative base period question.}
\label{tab:stara_abp_fps}
\begin{tabular}{|p{1.8cm}|p{4.5cm}|p{6.2cm}|}
\hline
\textbf{State} & \textbf{STARA Response} & \textbf{Verification Outcome} \\
\hline
Missouri &
\textit{True}. ``Section 288.501 specifically establishes an `alternate base period' for claimants who do not have sufficient wages in the regular base period. The statute states: `If a claimant does not have sufficient wages in the base period to be an insured worker...the individual's base period shall be the four most recently completed calendar quarters preceding the first day of the individual's benefit year. Such base period shall be known as the ``alternate base period''.' '' Cites Mo. Rev. Stat. \S~288.501. &
Correct (DOL Omission) — Missouri Code \S~288.501(1) explicitly provides: ``If a claimant does not have sufficient wages in the base period to be an insured worker, as those terms are defined in section 288.030, the individual's base period shall be the four most recently completed calendar quarters preceding the first day of the individual's benefit year. Such base period shall be known as the `alternate base period'.'' This is a valid alternative base period provision absent from the DOL compilation. The statute contains contingent provisions tied to the American Recovery and Reinvestment Act of 2009, but the core alternative base period authorization represents a legitimate provision missed by DOL. \\
\hline
Wyoming &
\textit{True}. ``Section 27-3-102 defines `base period' as `the first four (4) of the last five (5) completed calendar quarters immediately preceding the first day of an individual's benefit year or any other twelve (12) month period specified by commission regulation.' This statutory language explicitly authorizes the commission to establish alternative base period calculations beyond the standard regular base period.'' Cites Wyo. Stat. \S~27-3-102. &
Reasoning Error — Wyoming Statute \S~27-3-102(a)(i) grants the commission regulatory authority to define what constitutes a base period calculation but does not itself establish an alternative base period available to claimants who fail to qualify under the regular base period. An alternative base period requires an actual alternative calculation method available when the regular base period yields insufficient wages, not merely regulatory flexibility in defining the base period concept. Without evidence of commission regulations establishing such an alternative for claimants with insufficient wages, this statutory language does not demonstrate an alternative base period as defined by the evaluation standard. \\
\hline
\end{tabular}
\end{table}

\begin{table}[H]
\renewcommand{\arraystretch}{1.3}
\setlength{\tabcolsep}{3pt}
\centering
\small
\caption{Verification of Lexis+ AI apparent false positives on alternative base period question.}
\label{tab:lexis_abp_fps}
\begin{tabular}{|p{1.8cm}|p{4.5cm}|p{6.2cm}|}
\hline
\textbf{State} & \textbf{Lexis+ AI Response} & \textbf{Verification Outcome} \\
\hline
Missouri &
\textit{True}. ``Missouri uses an alternative base period for individuals who lack sufficient wages in the regular base period. This alternative base period includes the four most recently completed calendar quarters preceding the benefit year.'' Cites \S~288.501 R.S.Mo. &
Correct (DOL Omission) — As verified in STARA's analysis, Missouri Code \S~288.501(1) establishes a valid alternative base period provision missed in the DOL compilation. \\
\hline
Alabama &
\textit{True}. ``Alabama provides an alternative base period for individuals who do not qualify under the regular base period. The statute specifies that the last four completed calendar quarters may be used to determine eligibility if the individual lacks sufficient wages in the regular base period.'' Cites Code of Ala. \S~25-4-54, Code of Ala. \S~25-4-76. &
Reasoning Error — Neither cited provision establishes an alternative base period. Alabama Code \S~25-4-54 addresses benefit charge relief for employers under specific circumstances such as natural disasters and benefit adjustments, not alternative base period calculations for claimants. Alabama Code \S~25-4-76 addresses benefit calculation methods for maritime workers based on average quarterly earnings rather than highest two quarters, which is a different benefit computation method for a specific worker category, not an alternative base period available to claimants who fail monetary eligibility under the regular base period. \\
\hline
Tennessee &
\textit{True}. ``Tennessee allows for an alternative base period for individuals who lack sufficient wages in the regular base period. This alternative base period includes the last four completed calendar quarters immediately preceding the benefit year.'' Cites Tenn. Code Ann. \S~50-7-218. &
Reasoning Error — Tennessee Code \S~50-7-218 prevents overlapping base periods between consecutive benefit years by stipulating that ``if the first quarter of the last five (5) completed calendar quarters was included in the base period applicable to any individual's previous benefit year, the individual's base period shall be the last four (4) completed calendar quarters.'' This is an anti-overlap mechanism ensuring no quarter is used twice, not an alternative base period for claimants with insufficient wages. The provision also addresses extended base periods for workers' compensation recipients, which is distinct from an alternative base period. \\
\hline
Texas &
\textit{True}. ``Texas provides an alternative base period for individuals who are precluded from working due to a medically verifiable injury or illness. This alternative base period includes the first four calendar quarters of the five consecutive calendar quarters preceding the illness or injury.'' Cites Tex. Lab. Code \S~201.011. &
Reasoning Error (Extended Base Period) — Texas Labor Code \S~201.011(1)(B) provides an extended base period for individuals ``precluded because of a medically verifiable injury or illness from working during a major part of a calendar quarter,'' allowing use of ``the first four calendar quarters of the five consecutive calendar quarters preceding the calendar quarter in which the illness began or the injury occurred.'' This is an extended base period provision for individuals with workplace injuries or illnesses, not an alternative base period available to all claimants who fail monetary eligibility under the regular base period. The provision extends the lookback window for a specific circumstance rather than providing a general alternative calculation method. \\
\hline
\end{tabular}
\end{table}

\begin{table}[H]
\renewcommand{\arraystretch}{1.25}
\setlength{\tabcolsep}{3pt} 
\centering
\small
\caption{Representative Westlaw AI reasoning errors on alternative base period question (correct answer: False for all four states).}
\label{tab:westlaw_abp_errors}
\begin{tabular}{|p{1.8cm}|p{4cm}|p{6.7cm}|}
\hline
\textbf{State} & \textbf{Westlaw AI Response} & \textbf{Error Analysis} \\
\hline
Indiana &
\textit{True}. ``For individuals who received worker's compensation benefits for 52 weeks or less during the regular base period and, as a result, do not meet the wage credit requirements, the base period is redefined as the first four of the last five completed calendar quarters immediately preceding the last day the individual was able to work due to their injury.'' Cites IC 22-4-2-12.5. &
Reasoning Error (Extended Base Period) — Indiana Code \S~22-4-2-12.5 establishes an extended base period for workers' compensation recipients, not an alternative base period. The DOL report confirms this designation as ``EBP: Up to 4 quarters preceding the last day the individual was able to work.'' An extended base period extends the timeframe for wage calculation under specific circumstances (workplace injury with workers' compensation), whereas an alternative base period provides a different calculation method (typically the last four completed quarters) for claimants who fail general monetary eligibility. Westlaw AI conflates these distinct concepts. \\
\hline
Kentucky &
\textit{True}. ``If an individual lacks sufficient base-period wages due to a job-related injury and has received or was eligible to receive workers' compensation, an extended base period will be substituted for the current base period on a quarter-by-quarter basis as needed to establish a valid claim or to increase the benefit rate.'' Cites KRS \S~341.090. &
Reasoning Error (Extended Base Period) — Kentucky Revised Statute \S~341.090(1) explicitly establishes an ``extended base period'' for workers' compensation recipients, not an alternative base period. The statute uses the term ``extended base period'' throughout and limits availability to individuals who ``lack sufficient base-period wages because of a job-related injury'' and received workers' compensation. This extends the lookback window for specific circumstances rather than providing an alternative calculation method for claimants who fail general monetary eligibility. \\
\hline
Louisiana &
\textit{True}. ``Louisiana law provides for the use of an Alternative Base Period when a claimant fails to meet monetary eligibility under the regular base period and uses the last four completed calendar quarters.'' Cites LSA-R.S. 23:1472(4), LSA-R.S. 23:1600(5). &
Reasoning Error — Louisiana Revised Statutes \S~23:1472(4) defines ``base period'' as ``the first four of the last five complete calendar quarters immediately preceding the first day of an individual's benefit year.'' Section 23:1600(5) establishes monetary eligibility requirements but does not authorize an alternative base period. Westlaw AI provides no statutory language establishing an alternative for claimants with insufficient regular base period wages. The cited provisions define only the regular base period and eligibility conditions. \\
\hline
Mississippi &
\textit{True}. ``Mississippi law provides for the use of an alternative base period when a claimant fails monetary eligibility under the regular base period and uses the last four completed calendar quarters.'' Cites Miss. Code Ann. \S~71-5-11, Miss. Code Ann. \S~71-5-511, Miss. Code Ann. \S~71-5-513. &
Reasoning Error — Mississippi Code \S~71-5-11(A) defines ``base period'' as ``the first four (4) of the last five (5) completed calendar quarters immediately preceding the first day of an individual's benefit year.'' Sections 71-5-511 and 71-5-513 establish eligibility requirements and disqualification criteria without authorizing an alternative base period. None of these provisions create an alternative calculation method for claimants who fail monetary eligibility under the regular base period. \\
\hline
\end{tabular}
\end{table}

Beyond these illustrative examples, Westlaw AI's 12 apparent false positives exhibited consistent error patterns. In Alabama, Florida, and North Dakota, Westlaw AI cited base period definitions, benefit calculation provisions, or eligibility requirements without any language authorizing an alternative for claimants with insufficient regular base period wages. In Indiana and Kentucky, Westlaw AI mischaracterized extended base period provisions for workers' compensation recipients as alternative base periods, conflating these distinct concepts despite the DOL report explicitly designating them as extended base periods. In Louisiana, Mississippi, North Carolina, Pennsylvania, and Wyoming, Westlaw AI made entirely unsupported claims that states ``provide for'' or ``allow'' alternative base periods while citing only regular base period definitions and general eligibility requirements. In Tennessee and Texas, Westlaw AI cited anti-overlap provisions and injury-related extended base periods respectively, neither of which constitute alternative base periods available to claimants who fail general monetary eligibility. Across all 12 false positives, Westlaw AI consistently misinterpreted standard unemployment insurance provisions as alternative base period authorization, speculated about alternative mechanisms without statutory support, or conflated alternative base periods with extended base periods or other distinct legal concepts.

\section{Multi-Quarter Weekly Benefit Amount Method}
\label{app:multi-quarter}

The multi-quarter weekly benefit amount calculation question identifies states that compute unemployment benefits using wages from multiple quarters. States use various methods including high-quarter (most common), multi-quarter, annual-wage, and weekly-wage methods. The multi-quarter method aggregates wages across two or more quarters before applying the divisor, reflecting a broader employment pattern than single-quarter approaches. Westlaw AI generated 21 apparent false positives with six correct and 15 reasoning errors. Lexis+ AI produced five apparent false positives with two correct (Washington, Oregon) and three errors (New Hampshire, Wisconsin, West Virginia). The errors typically involved conflating eligibility requirements with benefit calculation formulas or misinterpreting multi-quarter eligibility thresholds as determining the benefit amount when the actual calculation used only the highest quarter.

Tables~\ref{tab:lexis_mq_errors} and~\ref{tab:westlaw_mq_errors} illustrate representative reasoning errors from Lexis+ AI and Westlaw AI, showing how both systems confused multi-quarter eligibility requirements with multi-quarter benefit calculation methods.
\begin{table}[H]
\renewcommand{\arraystretch}{1.25}
\setlength{\tabcolsep}{3pt} 
\centering
\small
\caption{Lexis+ AI reasoning errors on multi-quarter weekly benefit amount method (correct answer: False for both states).}
\label{tab:lexis_mq_errors}
\begin{tabular}{|p{2.5cm}|p{4cm}|p{6cm}|}
\hline
\textbf{State} & \textbf{Lexis+ AI Response} & \textbf{Error Analysis} \\
\hline
New Hampshire (False) &
\textit{True}. ``The weekly benefit amount is determined based on annual earnings, requiring earnings in at least two calendar quarters. This approach incorporates multiple quarters to calculate benefits, reflecting a more comprehensive employment pattern.'' Cites RSA 282-A:25. &
Reasoning error: The statute requires earnings in at least two quarters as an eligibility threshold, not as part of the benefit calculation. The actual benefit formula uses annual earnings divided by 26, which is an annual-wage method, not a multi-quarter aggregation method. Eligibility requirements do not determine calculation methodology. \\
\hline
Wisconsin (False) &
\textit{True}. ``Weekly benefits are calculated as 4\% of the base period wages paid during the quarter with the highest earnings, rounded down to the nearest dollar. This method incorporates multiple quarters to determine the base period wages.'' Cites Wis. Stat. \S~108.05. &
Reasoning error: The statute explicitly states benefits equal 4\% of wages ``paid during that quarter of the employee's base period in which the employee were paid the highest total wages.'' This is a standard high-quarter method (1/25 of highest quarter). The reference to ``base period wages'' does not change that only the highest quarter is used in the actual calculation. \\
\hline
\end{tabular}
\end{table}

\begin{table}[H]
\renewcommand{\arraystretch}{1.25}
\setlength{\tabcolsep}{3pt} 
\centering
\small
\caption{Westlaw AI reasoning errors on multi-quarter weekly benefit amount method (correct answer: False for both states).}
\label{tab:westlaw_mq_errors}
\begin{tabular}{|p{2.5cm}|p{4cm}|p{6cm}|}
\hline
\textbf{State} & \textbf{Westlaw AI Response} & \textbf{Error Analysis} \\
\hline
Iowa (False) &
\textit{True}. ``These provisions confirm that Iowa employs a multi-quarter method for calculating eligibility for unemployment benefits.'' Cites Iowa Code \S~96.4(4)(a) requiring wages at least 1.25 times highest quarter distributed across base period, and alternative base period using last four quarters. &
Reasoning error: Westlaw AI explicitly states these provisions determine ``eligibility for unemployment benefits'' rather than the benefit calculation method. The statute requires comparing total base-period wages to 1.25 times the highest quarter to determine if the claimant qualifies for benefits. This is a monetary eligibility threshold, not the formula for computing weekly benefit amounts. Westlaw AI conflates eligibility determination with benefit computation. \\
\hline
Kentucky (False) &
\textit{True}. ``This calculation method considers wages from multiple quarters (the base period) to determine eligibility and benefit amounts, rather than relying solely on a single quarter.'' Cites KRS 341.350(6) requiring highest quarter wages of at least \$1,500 and total base-period wages at least 1.5 times highest quarter, plus KRS 341.090 defining base period as four quarters. &
Reasoning error: These provisions establish monetary eligibility requirements, not the benefit calculation formula. The statute specifies minimum wage thresholds across quarters to qualify for benefits but does not indicate that wages from multiple quarters are aggregated or averaged in computing the weekly benefit amount. Westlaw AI states the method determines ``eligibility and benefit amounts'' but provides no statutory text showing multi-quarter aggregation in the actual benefit formula. \\
\hline
\end{tabular}
\end{table}

\section{Voluntary Contributions to Reduce Unemployment Tax Rates}
\label{app:voluntary-contributions}

The voluntary contributions question asks whether states permit employers to make voluntary payments to the unemployment compensation fund specifically to obtain lower contribution rates. The evaluation standard requires provisions to explicitly authorize employers to make voluntary contributions or payments and establish a clear connection between these payments and obtaining reduced or lower rates. Both elements must be present: authorization for voluntary payments and demonstration that such payments result in lower rates.

\begin{table}[H]
\renewcommand{\arraystretch}{1.25}
\setlength{\tabcolsep}{3pt} 
\centering
\small
\caption{Westlaw AI reasoning errors on voluntary contributions (correct answer: False for all four states).}
\label{tab:westlaw_vc_errors}
\begin{tabular}{|p{1.8cm}|p{4cm}|p{6.7cm}|}
\hline
\textbf{State} & \textbf{Westlaw AI Response} & \textbf{Error Analysis} \\
\hline
Alaska &
\textit{True}. ``Alaska Statute \S~23.20.170(c) establishes a standard rate of contributions for employment at 5.4 percent of wages paid, with reductions from the standard rate permitted under AS 23.20.280–23.20.310. This indicates that employers may take actions, such as voluntary contributions, to adjust their rates.'' Cites AS \S~23.20.170, AS \S~23.20.280. &
Reasoning error: The cited provisions establish Alaska's experience rating system and eligibility requirements for rate reductions based on compliance and experience, but contain no language authorizing voluntary contributions. Westlaw AI claims the general statement that ``reductions from the standard rate'' are permitted ``indicates that employers may take actions, such as voluntary contributions'' without any statutory text supporting this inference. The analysis conflates the existence of rate variation provisions with authorization for voluntary payments when these are entirely separate concepts. \\
\hline
Delaware &
\textit{True}. ``Employers may adjust their assessment rates based on their benefit wage ratio and other factors.'' Cites 19 Del.C. \S~3350. &
Reasoning error: Section 3350 describes how the Department calculates and administratively assigns rates based on employers' historical benefit wage ratios, not how employers can voluntarily make payments to reduce rates. The phrase ``adjust their assessment rates'' is misleading—the statute provides no mechanism for employers to proactively adjust rates through voluntary contributions. Westlaw AI fundamentally misinterprets rate calculation provisions as authorization for voluntary contribution mechanisms, conflating administrative rate determination with employer-initiated voluntary payments. \\
\hline
Tennessee &
\textit{True}. ``Employers can make voluntary contributions to improve their reserve ratio, which is used to calculate their UI tax rate. Voluntary contributions must be made within 120 days after the start of the rate year. These contributions are nonrefundable unless there is an error.'' Cites T.C.A. \S~50-7-403. &
Reasoning error: The cited statute contains absolutely no language about voluntary contributions, 120-day deadlines, or nonrefundability provisions. Tennessee Code \S~50-7-403 establishes only how reserve ratios are calculated (premiums paid minus benefits charged, divided by average taxable payroll) based on required payments through the normal experience rating system. Westlaw AI fabricates specific procedural details—including the 120-day deadline and nonrefundability rules—that do not exist anywhere in the cited provision, creating an entirely fictional voluntary contribution system without any statutory foundation. \\
\hline
Utah &
\textit{True}. ``When a successor acquires a business, the successor's contribution rate may be computed based on the combined experience of the predecessor and successor. This indicates that voluntary contributions can influence the rate calculation.'' Cites U.A.C. R994-303-106. &
Reasoning error: Utah Administrative Code Rule R994-303-106 addresses only successorship—how contribution rates are calculated when one business acquires another and experience rating accounts are transferred. This is a transfer of existing experience provisions, not a voluntary contribution mechanism. Westlaw AI claims successorship rate computation ``indicates that voluntary contributions can influence the rate calculation,'' fundamentally confusing the transfer of an acquired business's existing experience rating with authorization for employers to make voluntary additional payments to reduce their own rates. \\
\hline
\end{tabular}
\end{table}

\newpage

The DOL compilation originally identified 28 states permitting rate reduction through voluntary contributions. STARA identified 24 of these states, missing Louisiana, New Mexico, New York, and Michigan. Westlaw AI also identified 24 of the original 28, missing Kansas, West Virginia, New York, and Michigan. Lexis+ AI had substantially lower recall, identifying only eight of the original 28 states, all of which were also identified by the other systems.

Beyond the DOL's 28 states, Lexis+ AI identified no additional states. STARA identified one additional state (Oregon), where the provision mentions voluntary contributions but establishes no connection to rate reduction—a false positive stemming from incomplete statutory language. Westlaw AI identified 16 states not included in the DOL report, all of which proved to be reasoning errors through comprehensive verification. These errors followed systematic patterns: conflating experience rating systems with voluntary contribution authorization, misinterpreting payment method elections (contribution versus reimbursement) as voluntary rate reduction mechanisms, treating successorship rate calculations as evidence of voluntary contributions, and fabricating specific procedural details without statutory support. Table~\ref{tab:westlaw_vc_errors} presents four representative reasoning errors from Westlaw AI, demonstrating how the system consistently mischaracterized standard unemployment insurance provisions as voluntary contribution mechanisms.

Westlaw AI's errors exhibited consistent patterns across all 16 false positives. In Alabama and Florida, Westlaw AI cited purely procedural regulations governing protests and appeals of rate determinations as evidence of voluntary contribution authority. In Hawaii, Illinois, and Maryland, the system pointed to general rate calculation provisions and experience rating definitions, speculating without statutory support that these frameworks ``suggest'' or ``imply'' voluntary contribution mechanisms. In Mississippi and Nevada, Westlaw AI cited state-level reserve ratio definitions (measuring overall trust fund solvency) and incorrectly claimed these metrics could be improved through individual employer voluntary contributions. In Montana and New Hampshire, the system mischaracterized reimbursement payment elections (allowing certain employers to pay actual benefit costs instead of quarterly contributions) as evidence of ``flexibility'' in contribution methods that ``suggests'' voluntary payment authority. In Vermont and Virginia, Westlaw AI provided no specific statutory text and instead made completely speculative claims about ``the general framework'' supporting voluntary contributions, with Virginia's analysis acknowledging ``the provided statutes do not explicitly mention this 120-day deadline'' yet still concluding affirmatively based on fabricated assertions about ``general practice.'' Across all cases, Westlaw AI consistently conflated different statutory concepts—experience rating existence, rate calculation methodologies, payment method elections, successorship transfers, and appellate procedures—with authorization for voluntary contributions to reduce rates, despite the absence of any explicit voluntary contribution language in the cited provisions.

\end{document}